\documentclass[pageno]{jpaper}
\usepackage[subtle,title]{savetrees}
\usepackage{graphicx}
\usepackage{booktabs}
\usepackage{mathtools}
\usepackage{comment}
\usepackage{soul}
\usepackage{courier}
\usepackage{algorithm}
\usepackage{lipsum}
\usepackage{hyperref}
\hypersetup{colorlinks,linkcolor={blue},citecolor={blue},urlcolor={blue}}
\usepackage{multirow}
\usepackage{authblk}
\usepackage{amsmath,amssymb,amsfonts}
\usepackage{algorithmic}
\usepackage{textcomp}
\usepackage{xcolor}
\usepackage{xspace}
\usepackage{soul,color}  
\usepackage{breakurl}
\usepackage{ragged2e}
\usepackage{makecell}
\usepackage[figurename=Fig.,font={footnotesize,bf},labelfont=bf]{caption}
\usepackage{tikz}
\usepackage{fancyhdr}
\usepackage[normalem]{ulem}
\usepackage[sort,nocompress]{cite}
\usepackage{flushend}
\usepackage{letltxmacro}
\newcommand{\niparagraph}[1]{\vspace{0.5pt}\noindent\textbf{#1}}
\usepackage{enumitem} 
\newenvironment{enumpackedp}{%
\begin{enumerate}[label={(\arabic*)},noitemsep,nolistsep,leftmargin=*]
}
{%
\end{enumerate}
}
\newcommand{\bench}[1]{{\small{\textsf{#1}}}\xspace}
\newcommand{\benchtiny}[1]{{\tiny{\textsf{#1}}}\xspace}
\newenvironment{itempacked}{%
\begin{itemize}[noitemsep,nolistsep,leftmargin=*]
}
{%
\end{itemize}
}
\newcommand{\PCignore}[1]{}
\def\Snospace~{\S{}}

\renewcommand\figurename{Fig.}

\newcommand{\squishlist}{
 \begin{list}{$\bullet$}
  { \setlength{\itemsep}{0pt}
     \setlength{\parsep}{3pt}
     \setlength{\topsep}{3pt}
     \setlength{\partopsep}{0pt}
     \setlength{\leftmargin}{1.5em}
     \setlength{\labelwidth}{1em}
     \setlength{\labelsep}{0.5em} } }

\newcommand{\squishlisttwo}{
 \begin{list}{$\bullet$}
  { \setlength{\itemsep}{0pt}
     \setlength{\parsep}{0pt}
    \setlength{\topsep}{0pt}
    \setlength{\partopsep}{0pt}
    \setlength{\leftmargin}{2em}
    \setlength{\labelwidth}{1.5em}
    \setlength{\labelsep}{0.5em} } }

\newcommand{\squishend}{
  \end{list}  }

\newcommand{\rev}[1]{\textcolor{black}{#1}}

\newcommand\circled[1]{\tikz[baseline=(char.base)]{
            \node[shape=circle,fill=black,inner sep=0.5pt] (char) {\textcolor{white}{#1}};}}

\definecolor{plotred}{HTML}{941100}
\definecolor{plotgreen}{HTML}{008F00}
\definecolor{plotblue}{HTML}{005492}

\newcommand{\dataflow}{{\textsc{Flat}}\xspace}
\newcommand{\lshort}{{\sf L}\xspace}
\newcommand{\ashort}{{\sf A}\xspace}
\newcommand{\kshort}{{\sf K}\xspace}
\newcommand{\qshort}{{\sf Q}\xspace}
\newcommand{\vshort}{{\sf V}\xspace}
\newcommand{\oshort}{{\sf O}\xspace}
\newcommand{\FC}{{\sf FC}\xspace}
\newcommand{\CONV}{{\sf CONV}\xspace}
\newcommand{\GEMM}{{\sf GEMM}\xspace}
\newcommand{\flex}{{\textsc{Flex}}\xspace}
\newcommand{\naive}{{\sf Na\"ive}\xspace}

\hypersetup{breaklinks=true}
\newcommand{\asplossubmissionnumber}{81}
\usepackage[normalem]{ulem}
\usepackage{fdsymbol}
\definecolor{Gray}{gray}{0.9}
\newcommand{\CC}{\cellcolor{Gray}}
\usepackage{xcolor}

\PassOptionsToPackage{hyphens}{url}
\titlespacing\subsection{0pt}{1pt}{1pt}

\newcommand\blfootnote[1]{%
  \begingroup
  \renewcommand\thefootnote{}\footnote{#1}%
  \addtocounter{footnote}{-1}%
  \endgroup
}

\begin{document}

\fancypagestyle{firstpage}{
  \fancyhf{}
  \renewcommand{\headrulewidth}{1pt}
  \fancyhead[C]{\footnotesize{Appears in the Proceedings of the International Conference on Architectural Support for Programming Languages and Operating Systems, 2023}}
  \fancyfoot[C]{\thepage}
}

\title{\Large{\dataflow: An Optimized Dataflow for Mitigating Attention Bottlenecks}}
\author[]{%
Sheng-Chun Kao\quad{}\quad{}Suvinay Subramanian$^{\ast}$\quad{}\quad{}Gaurav Agrawal$^{\dagger{}1}$\quad{}\quad{}Amir Yazdanbakhsh$^\smallblackdiamond$\quad{}\quad{}Tushar Krishna\\
\small{Georgia Institute of Technology\quad{}\quad{}$^{\dagger}$Microsoft\quad{}\quad{}$^{\ast}$Google\quad{}\quad{}$^\smallblackdiamond$Google Research, Brain Team}\\\vspace{0.3cm}
\footnotesize
{\texttt{\href{mailto:felix@gatech.edu}{felix@gatech.edu}, \href{mailto:suvinay@google.com}{suvinay@google.com}, \href{mailto:gaagrawal@microsoft.com}{gaagrawal@microsoft.com}}
\\
\texttt{\href{mailto:ayazdan@google.com}{ayazdan@google.com},
\href{mailto:tushar@ece.gatech.edu}{tushar@ece.gatech.edu}
}}
}
\date{}
{
\let\clearpage\relax
\maketitle
}
\pagenumbering{arabic} 
\thispagestyle{firstpage}
\pagestyle{plain}
\begin{abstract}
Attention mechanisms, primarily designed to capture pairwise correlations between words, have become the backbone of machine learning, expanding beyond natural language processing into other domains.
This growth in adaptation comes at the cost of prohibitively large memory requirements and computational complexity, especially at higher number of input elements.
This limitation is due to inherently limited data reuse opportunities and quadratic growth in memory footprints, leading to severe memory-boundedness and limited scalability of input elements.
This work addresses these challenges by devising a tailored dataflow optimization, called \dataflow, for attention mechanisms without altering their functionality.
This dataflow processes costly attention operations through a unique fusion mechanism, transforming the memory footprint quadratic growth to merely a linear one.
To realize the full potential of this bespoke mechanism, we propose a tiling approach to enhance the data reuse across attention operations.
Our method both mitigates the off-chip bandwidth bottleneck as well as reduces the on-chip memory requirement.
\dataflow delivers 1.94$\times$ (1.76$\times$) speedup and 49\% and (42\%) of energy savings compared to the state-of-the-art Edge (Cloud) accelerators with no customized dataflow optimization.
\rev{When on-chip resources are scarce (20\,KB-200\,KB), \dataflow yields, on average, 1.5$\times$ end-to-end latency reduction across a diverse range of conventional attention-based models with input sequence lengths ranging from 512-token to 64K-token.}
Our evaluations demonstrate that state-of-the-art DNN dataflows applied to attention operations reach the efficiency limit for inputs above 512 elements.
In contrast, \dataflow unblocks transformer models for inputs with up to 64K elements.
\blfootnote{$^1$~Work done when Gaurav Agrawal was at Google.}
\end{abstract}
\section{Introduction}
\label{sec:intro}
Attention mechanisms, the key building block of transformer models, have enabled state-of-the-art results across a wide range of machine learning (ML) tasks---from natural language processing (NLP)~\cite{transformer, TrXL, xlm_model}, to object detection~\cite{zhu2020deformable,bhattacharyya2021self,sun2020transtrack}, image classification~\cite{vision_longformer, levit, cvt, longshort_transformer}, image generation~\cite{imagegpt,parmar2018image,esser2020taming}, and music synthesis~\cite{huang2018music,hsiao2021compound}.
This exponential growth of transformer models are expected to serve as the foundation of a new bread of machine learning models in the upcoming years.
A key attribute of attention-based models is the \textit{sequence length ($N$)} defining the number of input elements for which a pairwise correlation scores is computed.
Intuitively, increasing sequence length enables the attention-based models to better capture the context of input sentences or the relation between image segments.
The demand for leveraging long-sequence (e.g. $N=8K$ to $N=69K$) attention-based models has already emerged in ML community~\cite{long_range_arena}, beyond natural language understanding~\cite{compressive_transformer} into protein folding~\cite{performer} and text summarization~\cite{kitaev2020reformer} and audio generation~\cite{liu2018generating}.
\rev{Employing long sequences is pivotal in these algorithms because the property of input emerges from the global context. For example, two proteins may look identical if we examine identical sequence fragments, but when the entire sequence is considered, the differences in their function arise.
We observe an analogous phenomenon in text summarization, where context can drastically alter the meaning of the selected text subset. In that instance, the subset represents a shorter sequence while the entire context refers to the full-length one.}

Compared to existing neural network accelerators~\cite{eyeriss_v2, chen2016eyeriss,du2015shidiannao,simba,qin2020sigma,yazdanbakhsh2018ganax}, architecting accelerators for attention-based models poses different design challenges, attributed to their soaring demand for on-chip memory and compute complexities.
Recent accelerators for attention-based models~\cite{a3,elsa} have mainly relied on algorithmic optimizations, often with negative repercussion on model accuracy. 
Algorithmic techniques in practice include sparsification or compression~\cite{wang2020linformer,katharopoulos2020transformers,choromanski2020masked,performer,shen2018bi,parmar2018image,qiu2019blockwise,child2019generating,beltagy2020longformer,correia2019adaptively,kitaev2020reformer,roy2021efficient,tay2020synthesizer,TrXL,compressive_transformer,leopard,sprint:micro} and/or leveraging lossy approximation~\cite{elsa,a3,wang2021spatten}.

In this work, we identify that the conventional dataflow/mapping methods for \CONV and \FC layers~\cite{nvdla,du2015shidiannao,tpu,chen2016eyeriss} are inadequate for attention layers. 
This is because the main operators within attention layers exhibit distinct compute and memory characteristics posing notable bottlenecks on off-chip memory bandwidth compared to \CONV and \FC.
We identify the following challenges in devising dataflow optimizations for attention layers:
\begin{enumpackedp}
\item \niparagraph{Significantly low operational intensity.} Inherently low data reuse in activation-activation operators significantly reduces the operational density of such operators in attention layers. This inherently low operational density subsequently makes the activation-activation operators \textit{fundamentally} memory-bound.
While prior work on intra-operator dataflow optimization, such as loop transformation and scheduling techniques~\cite{nvdla,eyeriss_v2,du2015shidiannao,timeloop,maestro, synthesislectures_orchestration}, targets \CONV and batched \FC operators by leveraging the ample intrinsic data reuse, which are not well-suited for activation-activation operators lacking data reuse opportunity.
\item \niparagraph{Complex many-to-many operators.} The main attention operators have many-to-many relation, obscuring opportunities to use operator fusion~\cite{dnnfusion} in attention operators. That is because operator fusion in conventional ML compilers~\cite{autotvm,ragan2013halide,tensorflow_xla} mainly target operations such as \CONV and \FC with one-to-one (i.e. element-wise) relation.
\item \niparagraph{Prohibitively large intermediate tensors.} The size of intermediate tensors in attention layers grows quadratically\textemdash{}$\mathcal{O}(N^2)$)~\cite{transformer,TrXL, xlm_model,wang2020linformer,katharopoulos2020transformers,choromanski2020masked,performer}\textemdash{}with the sequence length. This quadratic growth imposes a significant pressure on on-chip memory capacity and prohibits opportunities to store the intermediate results on-chip and improve the compute utilization, a common practice in CNN accelerators~\cite{chen2016eyeriss}.
\end{enumpackedp}

\noindent{}This paper fundamentally tackles the challenges associated with attention layers by devising a first in its class \textit{many-to-many inter-operator} dataflow optimization mechanism, called \textbf{\underline{F}}used \textbf{\underline{L}}ogit \textbf{\underline{A}}ttention  \textbf{\underline{T}}iling.
This optimization particularly fuses multiple many-to-many tensor operator, while systematically preserving their inter-operator data dependencies, leading to a significant reduction on off-chip memory bandwidth pressure.
In addition, to fully realizing the performance benefit of this inter-operator fusing mechanism, \dataflow performs a new tiling approach across the fused operators.
This tiling enables efficient staging of quadratically growing intermediate tensors of attention operations on tight-budgeted on-chip memories, leading to higher performance and energy savings and elevates the scalability of transformer models up to 64K inputs.
These benefits are unlocked with only modest hardware changes, integrating into a platform deployable on off-the-shelf DNN accelerators.
In summary, this paper presents the following specific contributions for attention-based models:
\begin{itempacked}
\item We systematically study the operational intensity of different operators within attention layers and characterize the fundamental roadblocks imposed by limited hardware resources to improve the overall realized performance of attention accelerators (\autoref{sec:motivation}).
\item Based on the resulting findings, we explore fusion opportunities between different operators in attention layers and justify our proposed many-to-many inter-operator fusion (\autoref{sec:flat}).
While beneficial, this fusion inflicts a fundamental challenge of preserving the inter-operator data dependencies, imposed by the softmax operation.
To address, we expound our tailored dataflow optimization approach for attention layers, enabling higher data reuse of the quadratically growing intermediate attention tensors from low-capacity but high-bandwidth on-chip memory. We show that this dataflow optimization efficiently mitigates the pressure on off-chip memory bandwidth, leading to a higher performance and energy efficiency in accelerators (\autoref{sec:dataflow_implementation}).
\item We develop a map-space exploration framework to efficiently search for optimal loop orders across fused operators and tiling sizes. This framework optimizes performance metrics of interest subject to different hardware resource constraints, such as number of processing elements and on-chip memory capacity (\autoref{sec:analytical_model}).
\end{itempacked}

\noindent{}We evaluate \dataflow on a variety of Attention-based models, including \bench{BERT}~\cite{transformer}, \bench{TrXL}~\cite{xlm_model}, \bench{FlauBERT}~\cite{le2019flaubert}, \bench{T5}~\cite{t5_model}, and \bench{XLM}~\cite{xlm_model}, for both Edge and Cloud accelerators.
Compared to a range of state-of-the-art dataflow optimizers, \dataflow delivers 1.75$\times$ and 1.65$\times$ speedup and 44$\%$ and 55$\%$ energy savings for recent Edge and Cloud accelerators, respectively.
When on-chip resources are scarce (in the order of 10KB-100KB), \dataflow yields 1.5x end-to-end latency reduction and 1.4x end-to-end energy savings for the Attention-models with conventionally-sized input sequence length (512-token).
Furthermore, our results show that while the conventional DNN dataflow optimizers for attention operations bumps up against the efficiency limit for inputs above 512 tokens, our dataflow optimization tailored for attention operations unblocks the scalability of transformer models for significantly larger input sizes, up to 64K tokens.
\section{Background on Attention-Based Models}
\label{sec:background}

\begin{figure}
\begin{center}
\includegraphics[width=0.99\linewidth]{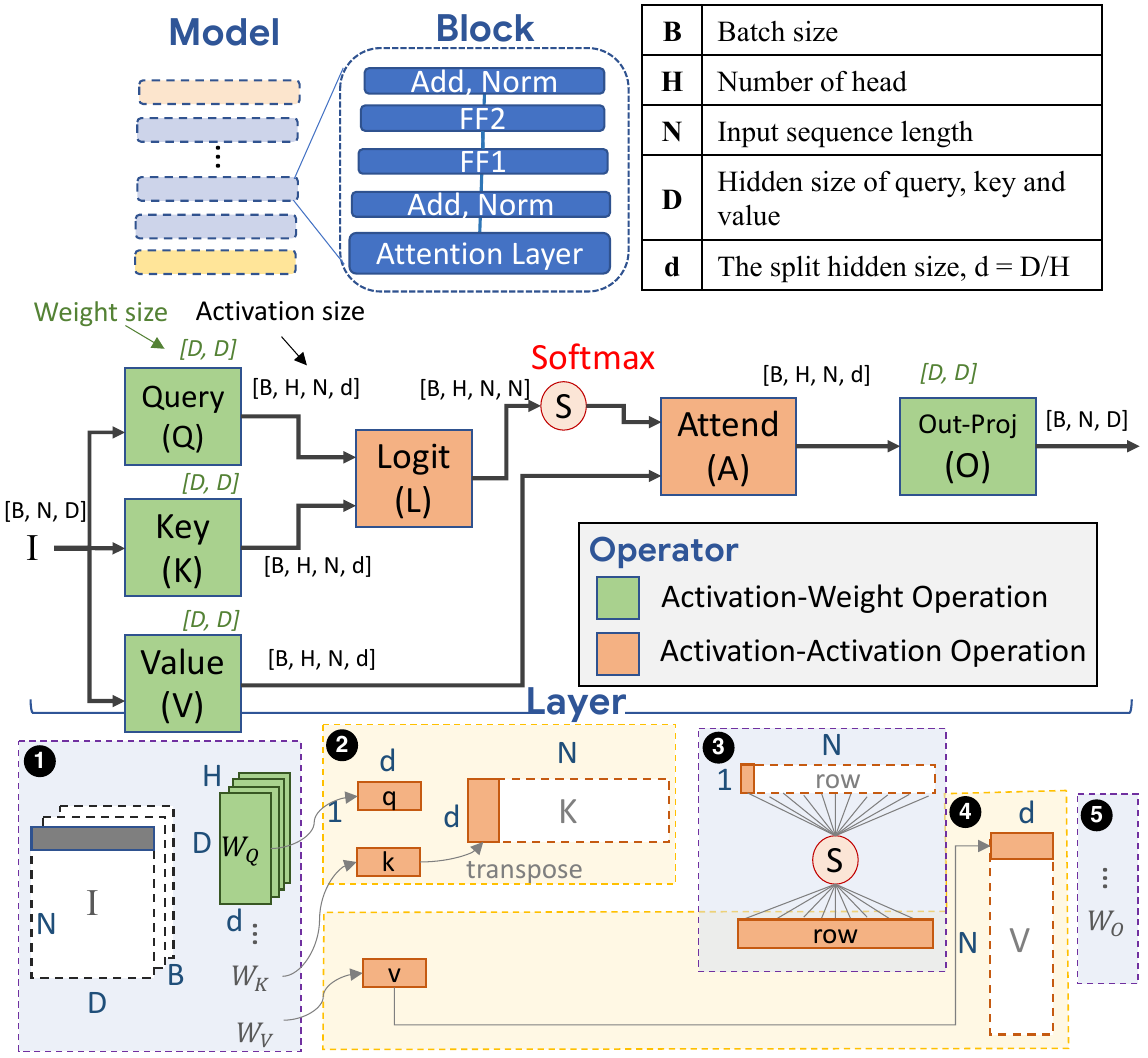}
\end{center}
\caption{The structure of attention-based models. Green matrix notation shows the size of weight tensors; black matrix notation shows the size of activation tensors; Softmax is applied on output of Logit.}
\label{fig:compute_flow}
\end{figure}

\niparagraph{Terminology.}
Transformer models~\cite{transformer,le2019flaubert, sun2020mobilebert} generally share similar architectures. In a top-down view (\autoref{fig:compute_flow}), an attention-based \textit{``model''} comprises multiple (often identically parameterized) attention \textit{``blocks''}\footnote{Models may include other blocks: an embedding block with positional encoding and masking, and a few task-specific \FC or \CONV layers.}.
An attention block comprises multiple \textit{``layers''}: an attention layer, a normalization layer, followed by multiple (typically two) fully connected layers. Finally, each layer comprises one or more \textit{``operations''} or \textit{``operators''}.

\niparagraph{Computation operators.}
The attention mechanism measures how closely two \emph{tokens} are related in an input \emph{sequence}.
Each token in the input sequence is represented as a vector of dimension $D$, each sequence has $N$ tokens, and an input batch to an attention layer comprises a batch of $B$ sequences; thus the input to the attention layer is a tensor of dimension $[B,N,D]$. 
\autoref{fig:compute_flow} (bottom) highlights the flow of a single token vector (of dimension $D$) through the attention mechanism. Step \circled{1}, three vectors are derived for each token's vector in the input: called Key (K), Query (Q), Value (V).
This is achieved by multiplying the input tensor with learnable weight matrices. Attention mechanisms often use multiple \emph{heads} to generate $H$ such K, Q, V vectors for each token vector in the input. Thus each of K, Q, V generates a tensor of dimension $[B,H,N,d]$. Step \circled{2}, we compute the \emph{logits score} (L) which captures how strongly each token is related to each other token in the sequence.
This is done by a ($d$-dim) dot product of each vector of the key with the corresponding vector of the query. Each dot product yields a single scalar score, but this score is computed for each token against all other tokens in the sequence yielding an output tensor of dimension $[B,H,N,N]$.
Step \circled{3}, the logits scores then needs to be normalized. While there are several normalization functions, they all share key traits.
The normalization is carried out across a \emph{row} of $N$ logits scores in each sequence.
To generate a row of $N$ normalized scores, the normalization operator reads $N$ input scores, performs a reduction of these $N$ scores, and scales each of the $N$ input scores by the reduced value. We use softmax since it is the most commonly used normalization function.
Step \circled{4}, involves performing a weighted sum of the value vectors with the corresponding weights from the logits, which yields an output tensor $[B,H,N,d]$. Finally, step \circled{5} concatenates the attention outputs from the $H$ heads and with its weight matrix computes a output tensor $[B,N,D]$. These complete an attention layer.

This can be represented succinctly as the computational graph in \autoref{fig:compute_flow}, comprising the following operators: i) Query (\qshort), Key (\kshort), and Value (\vshort) operators that perform a projection of the input tensor, ii) Logit (\lshort) and Attend (\ashort) operators that compute the logits scores and weighted-sum of values respectively, and iii) Output (\oshort) operator that performs an output projection.  We categorize them into two: (i) activation-weight operators (\qshort, \kshort, \vshort, \oshort), which operate on activation tensors (from previous operators) and weight tensors (model parameters) and perform a GEMM computation as conventional fully connected operators (FCs), and ii) activation-activation operators (\lshort, \ashort), which operate on two activations from different previous operators and perform a GEMM computation. The \lshort and \ashort operators often dominate the latency and power consumption while running the model~\cite{a3}, and even more so at long sequence lengths, as shown in our evaluations (\autoref{fig:latency_ratio}).

\subsection{DNN Accelerators - Performance Considerations}
\label{sec:dnn_accelerators}

We consider spatial DNN accelerators~\cite{chen2016eyeriss, synthesislectures_orchestration,edgetpu} in this work (\autoref{sec:exp_MSE} provides more details).
We discuss the key factors that determine the realized performance when running a DNN model on an accelerator with specific hardware resources (PEs, on-chip memory size, and off-chip memory BW).

\niparagraph{Operational intensity | Roofline performance.}
\label{sec:op_intensity}
Operational intensity is a proxy metric to gauge the maximum possible performance of an \emph{individual} operator given a set of hardware resources.
The operational intensity of an operator is defined as the number of arithmetic operations divided by the number of memory accesses. A lower operational intensity implies an operator has fewer opportunities for data reuse and is more likely to be memory bandwidth(BW)-bounded. This directly decides the roofline (or best achievable) performance of the operator on the underlying accelerator.
\begin{align}
\hskip\parindent & \begin{gathered}
    \mathcal{I} = \frac{\#\ \mathrm{of}\ \mathrm{Operations}}{\#\  \mathrm{of}\ \mathrm{Memory}\ \mathrm{Accesses}}
    \end{gathered} 
\label{formula:op_intensity}
\end{align}

\niparagraph{Dataflow | Realized performance.}
\label{sec:dataflow_def}
Dataflow refers to the mechanisms to stage data from the off-chip memory through the on-chip memory hierarchy to the compute PEs, over space and time~\cite{synthesislectures_orchestration}.
It determines the actual achieved performance.
Since memory access is often the bottleneck in executing DNN operators~\cite{synthesislectures_efficient}, the dataflow exposes data reuse opportunities across operands that can be exploited in hardware via buffering and data forwarding/broadcast~\cite{synthesislectures_orchestration}.
Formally, the \textit{dataflow} encompasses:
(i) tiling (how tensors are sliced, stored and fetched across the memory hierarchy), (ii) compute order (order in which loop iterations are performed), 
and (iii) parallelism (how compute is mapped across PEs spatially). 
The dataflow along with specific tile sizes is often called a \textit{mapping}~\cite{maestro,synthesislectures_orchestration}.
\section{Challenges with Running Attention Layers}
\label{sec:motivation}
\subsection{Challenge 1: Low Operational Intensity of L/A}
\label{sec:challenge_opintensity}

\niparagraph{Activation-Weight operators (\qshort/\kshort/\vshort/\oshort).}
Following the notation in \autoref{fig:compute_flow}, the number of operations in these operators is $\mathcal{O}(BND^{2})$. 
The number of memory accesses for the input (activations), weight (parameters), and output (activations) tensors are $\mathcal{O}(BND)$, $\mathcal{O}(D^{2})$, $\mathcal{O}(BND)$, respectively. Therefore the operational intensity is $\mathcal{O}(\frac{BND^{2}}{BND+D^{2}+BND})$. 
We see that increasing the batch size (B) can increase the operational intensity---the same weight value can be \emph{reused} by multiple activations, leading to lower BW pressure. 
This is 
a typical technique used in activation-weight operators (e.g., \CONV and \FC, the staple in most DNN models) as it makes better use of the scarce memory bandwidth in accelerators and enables higher utilization of the provisioned compute FLOPs, leading to improved throughput.
The specific mechanism to exploit the operational intensity is called dataflow~\cite{chen2016eyeriss, synthesislectures_orchestration}.

\begin{figure}
\begin{center}
\includegraphics[width=0.99\linewidth]{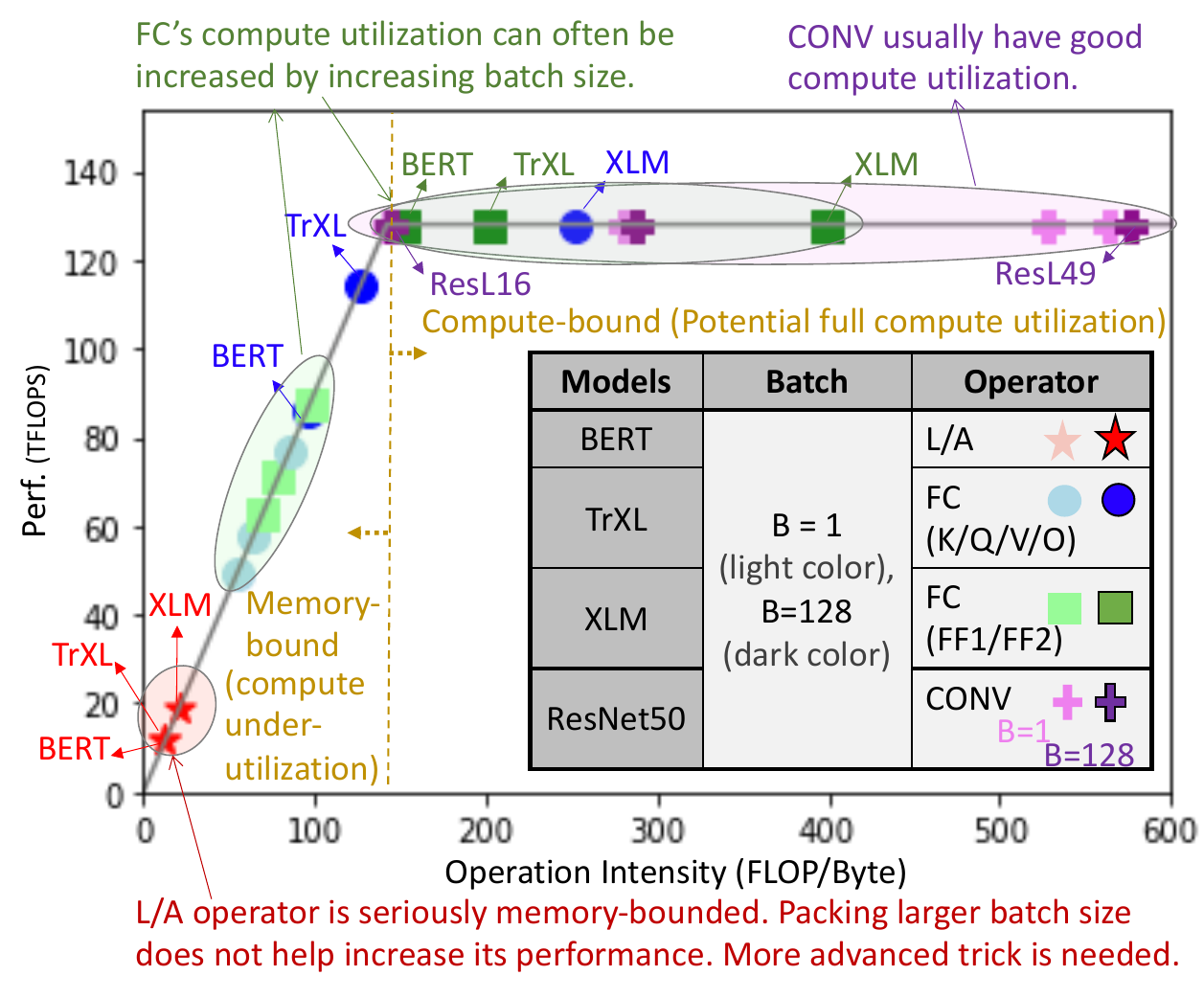}
\end{center}
\caption{Roofline analysis on TPU-v3~\cite{tpuv3} for operators in  BERT(-base)~\cite{transformer}, TrXL(-large)~\cite{TrXL}, and XLM (xlm-mem-en)~\cite{xlm_model}, and ResNet50~\cite{resnet} using sequence length = 512.}
\label{fig:roofline_quant}
\end{figure}

\niparagraph{Activation-Activation operators (\lshort/\ashort).}
\label{sec:L/A operators}
For \lshort and \ashort operators, the number of operations is $\mathcal{O}(BN^{2}D)$. The number of memory access for the two input-activations and the output-activations are $\mathcal{O}(BND)$, $\mathcal{O}(BND)$, $\mathcal{O}(BN^{2})$, respectively. Therefore the operational intensity is $\mathcal{O}(\frac{BN^{2}D}{2BND+BN^{2}})$.
Embedding size (D) is decided by the model, and sequence length (N) is decided by the application. Furthermore, multi-head attention is an often-used variant of the attention mechanism: it leads to higher accuracy in many tasks~\cite{transformer}. It splits the output of the \qshort/\kshort/\vshort operator along a hidden dimension, reshaping it from size [N, D] to [H, N, d], where d=D/H. The operational intensity of \lshort, \ashort becomes $\mathcal{O}(\frac{BN^{2}D}{2BND+BHN^{2}})$. For these operators, one can not simply increase the batch size to increase the operational intensity.

\begin{figure}
\begin{center}
\includegraphics[width=0.99\linewidth]{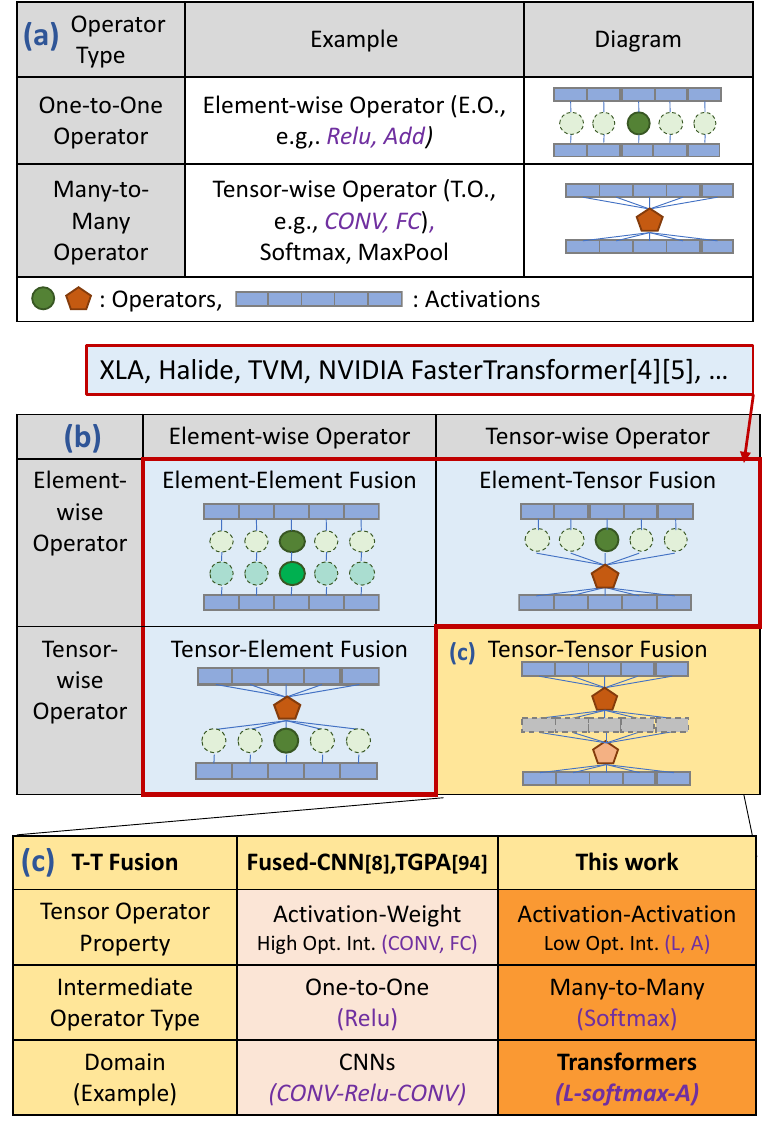}
\end{center}
\caption{(a) Type of Operator, (b) Type of Operator-Fusion, and (c) Type of Tensor-Tensor Fusion.}
\label{fig:classify_table}
\end{figure}

\niparagraph{Roofline analysis.}
To quantitatively demonstrate the effect of operation intensity across different operators, we show the roofline analysis of operators of three common attention-based models~\cite{transformer, TrXL, xlm_model} and a widely used CNN network ResNet50~\cite{resnet} on TPU-v3~\cite{tpuv3} in \autoref{fig:roofline_quant}.
We can see that \CONV operators lie in the compute-bound region. \FC operators scatter across both memory and compute bound region; however, with the increase in batch size, their operation intensity increases and can become compute-bound. This demonstrates why batching is a popular technique for \FC layers.
In contrast, \lshort and \ashort operators sit at memory-bound and low-performance region, and batch size increase is not effective in these operators (\autoref{sec:L/A operators}).

\setlength{\fboxsep}{2pt}
\noindent\fbox{%
\begin{minipage}{0.475\textwidth}{
    In summary, low operational intensity of the individual \lshort/\ashort operators makes them fundamentally memory-bound, and any dataflow/mapping exploration at the individual operator level cannot further improve performance.
    }
\end{minipage}}

\subsection{Challenge 2: Complexity of Op Fusion for L/A}
\label{sec:challenge_dependency}
Given the low operational intensity for \lshort/\ashort operators, fusion is an attractive technique to stage the intermediate tensor data on-chip and leverage the higher on-chip memory bandwidth.
Operator fusion is an optimization that schedules back-to-back operators together such that the producer's output  directly feeds the consumer, thus avoiding materialization of full intermediate tensor in memory. 
By avoiding off-chip data movement of the intermediate tensor, we can use the higher on-chip bandwidth to enable improved performance for the fused operator (as opposed to executing the operators individually).

When exploring operation fusion opportunities, we can either fuse among Element-wise Operators (E.O.) or Tensor-wise Operators (T.O.), as shown in \autoref{fig:classify_table}. Element-Element fusion (E-E) is the simplest fusion optimization. With increased interest in operation fusion, more ML compilers/frameworks today support Tensor-Element (T-E) or Element-Tensor (E-T) fusion~\cite{autotvm, baghdadi2019tiramisu, TACO, tensorflow_xla, ragan2013halide} 
where \textit{MatMul operators} (i.e., \CONV or \FC) are often fused with element-wise  operators (such as ReLu or Add), reshapes, or shuffling operators~\cite{dnnfusion}

However, Tensor-Tensor Fusion (T-T) is not done automatically. The key reason is that T.O. is a many-to-many operator (\autoref{fig:classify_table}(a)). While it is often straightforward and a simple  engineering exercise to fuse a T.O. with one-to-one operator like E.O., how to fuse many-to-many with other many-to-many and whether it is beneficial to fuse them is still a research question~\cite{dnnfusion}.
Indeed, DNNFusion~\cite{dnnfusion} which studied T-T as recently as PLDI 2021, reported it to be either too complicated or unprofitable.
The key reason is that the additional complexity to maintain dependence and stage data (grey intermediate data as shown in \autoref{fig:classify_table}(b)) could end up negatively impacting register and cache usage~\cite{dnnfusion}.

Some previous research papers~\cite{fused_cnn, wei2018tgpa} have discussed Tensor-Fusion in CNNs with the fusion pattern T-(one-to-one)-T such as CONV-Relu-CONV and shown huge potential gain with a well-designed inter-operator dataflow for Tensor-Fusion. This is highlighted in \autoref{fig:classify_table}(c). However, Tensor-Fusion for attention-based models has not been explored to date, to the best of our knowledge. The complexity of its fusion pattern, T-(many-to-many)-T, makes it much more challenging than for CNNs.

\setlength{\fboxsep}{2pt}
\noindent\fbox{%
\begin{minipage}{0.47\textwidth}{
    To address this challenge, we design a specialized inter-operator dataflow that not only considers the data dependency of two large tensor operators but also tackles the additional complex data dependency incurred by the many-to-many intermediate activation (\autoref{sec:constraint_dependency}).
    }
\end{minipage}}
\begin{figure}
\begin{center}
\includegraphics[width=0.97\linewidth]{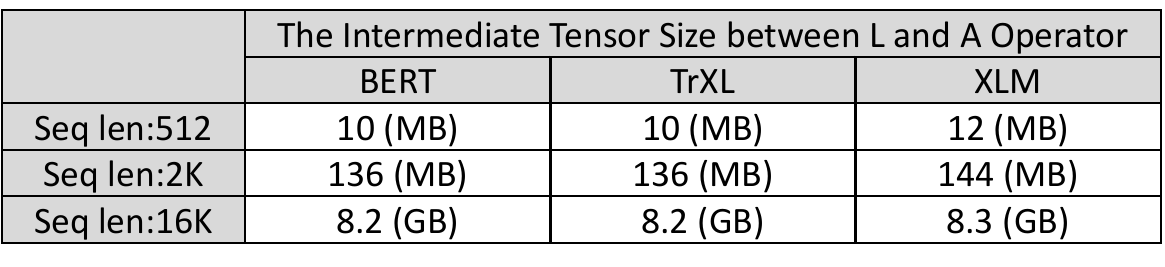}
\end{center}
\caption{Intermediate tensor size between \lshort and \ashort operators in BERT(-base)~\cite{transformer}, TrXL(-large)~\cite{TrXL}, and XLM(xlm-mem-en)~\cite{xlm_model}.}
\label{fig:roofline}
\end{figure}

\subsection{Challenge 3: Tensor Footprint of L/A}
\label{sec:challenge_memfootprint}
There is one other challenge that is unique to \lshort/\ashort when considering Tensor-Fusion---namely a \textit{quadratic} intermediate tensor footprint.
From \autoref{fig:compute_flow} we can calculate the intermediate tensor between L and A operators has size $\mathcal{O}(BHN^2)$ (M-Gran in \autoref{table:l3_size_of_example}).
This footprint grows quadratically with sequence length, and exceeds 8GB (exceeding the viable on-chip memory in many data-center class accelerators~\cite{tpuv3, tesla2018v100}) beyond 
sequence lengths of 16K (\autoref{fig:roofline}).
As NLP tasks with larger sequence lengths become popular~\cite{kitaev2020reformer, compressive_transformer,performer}, the technique of keeping the entire intermediate tensor on-chip is not scalable.
\setlength{\fboxsep}{2pt}
\noindent\fbox{%
\begin{minipage}{0.47\textwidth}{
    To address this, we propose a tiling technique for our fused operator that enables controlling the active memory footprint based on the on-chip memory constraint.
    }
\end{minipage}}
\begin{figure}
\begin{center}
\includegraphics[width=0.99\linewidth]{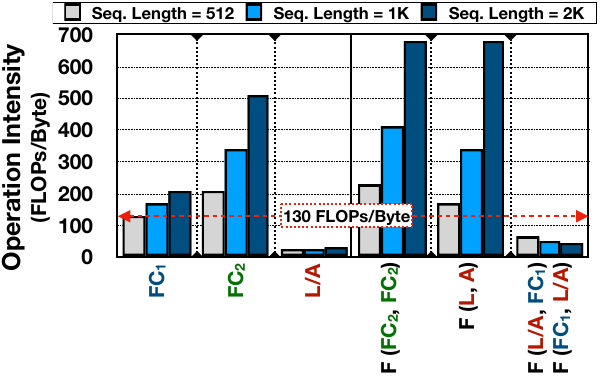}
\end{center}
\caption{Potential of Tensor-Tensor Fusion. The operation intensity ($\mathcal{I}$) of single and fused operators in attention layers in TrXL(-large)~\cite{TrXL} using batch size=1.
The notations are from \autoref{fig:compute_flow}. F(X, Y) implies fusion between X and Y operator.
The red dashed bar indicates the minimum operation intensity for the operator to become compute-bound.
FC$_1$ and FC$_2$ indicate operator K/Q/V/O and FF1/FF2, respectively. F(L/A, FC$_1$) specifies fusion between ``A'' and ``O'', whereas F(FC$_1$, L/A) expresses the fusion between ``Q-L'', ``V-A'', and ``K-L''.}
\label{fig:fused_quant}
\end{figure}

\section{\dataflow Dataflow Concept}
\label{sec:flat}
We design a specialized dataflow strategy, \textbf{\underline{F}}used \textbf{\underline{L}}ogit \textbf{\underline{A}}ttention  \textbf{\underline{T}}iling (\dataflow), targeting the two memory BW-bound operators in the attention layer, \lshort and \ashort.
\dataflow includes both intra-operator dataflow and a specialized inter-operator dataflow, executing \lshort and \ashort in concert. 

\subsection{Identifying Tensor Fusion Opportunity}
\label{sec:fusion}
\autoref{fig:fused_quant} plots the 
operation intensity ($\mathcal{I}$) of single and fused operators in attention layers of an attention-based model~\cite{TrXL}. The dotted line marks the operation intensity threshold (ridge point) from memory to compute boundedness in TPU-v3~\cite{tpuv3}.
We observe that for FC-based operators (\kshort/\qshort/\vshort/\oshort), the operational intensity is sufficient to be compute-bound, while for L/A it is low (as we had also observed via \autoref{fig:roofline_quant}).
However, after fusing \lshort and \ashort (f(\lshort, \ashort)), the effective operational intensity (of the fused operator) is higher.
This motivates us to explore \lshort and \ashort fusion.

\niparagraph{Why not fuse other operators?}
We did not fuse other operator pairs such as f(\qshort, \lshort), f(\ashort, \oshort), or f(\vshort, \ashort), for three reasons. (1) The operational intensity is often sufficient and can be increased by leveraging batch size to reach compute-bound (\autoref{fig:roofline_quant}). (2) Fusing two FCs (f(\FC, \FC)) can achieve higher operational intensity; however since the operator is already compute-bound, there is not much value in leveraging fusion (and the additional complexity).
(3) We often need finer-granularity dataflow schemes to fit fused operator tensors on-chip; however fusing two activation-weight computation (f(\FC, \FC)) can trade-off (weight) reuse opportunity and may reduce actual achievable performance (\autoref{sec:gran_tradeoff}).

\niparagraph{Why not fuse multiple operators?}
We did not fuse multiple operators such as f(\lshort, \ashort, \oshort) or f(\kshort, \lshort, \ashort) for two reasons.
(1) Fusing \lshort/\ashort with \FC such as f(\ashort, \oshort) or f(\kshort, \lshort) can drop the potential performance of FCs compared to their single operator performance (\autoref{fig:fused_quant}). (2) The more operators we fuse, the more data we need to stage partially on-chip. Since the on-chip memory is often extremely limited, we need to execute the fused operators at a much finer granularity, which may lead to a degradation in achievable performance (\autoref{sec:R-gran_tradeoff}). With these analyses, we decide to fuse only \lshort and \ashort.
\subsection{Challenges with Tensor Fusion Implementation}
\label{sec:fusion_challenges}

Fusing \lshort and \ashort operators introduces two key challenges that we discuss here. \autoref{sec:dataflow_implementation} presents implementation details.

\niparagraph{Challenge 1: Respecting data dependencies across operators.}
Fusing \lshort and \ashort causes its unique challenge of data dependency owing to the many-to-many Softmax operation between them (\autoref{sec:challenge_dependency}).
Softmax requires a reduction along a specific dimension of the tensor before scaling individual elements. Arbitrary inter-loop tiling as employed by prior \CONV/\FC fusion techniques~\cite{fused_cnn, wei2018tgpa} violates this data dependency constraint.

\begin{figure*}[ht]
\begin{center}
\includegraphics[width=0.99\linewidth]{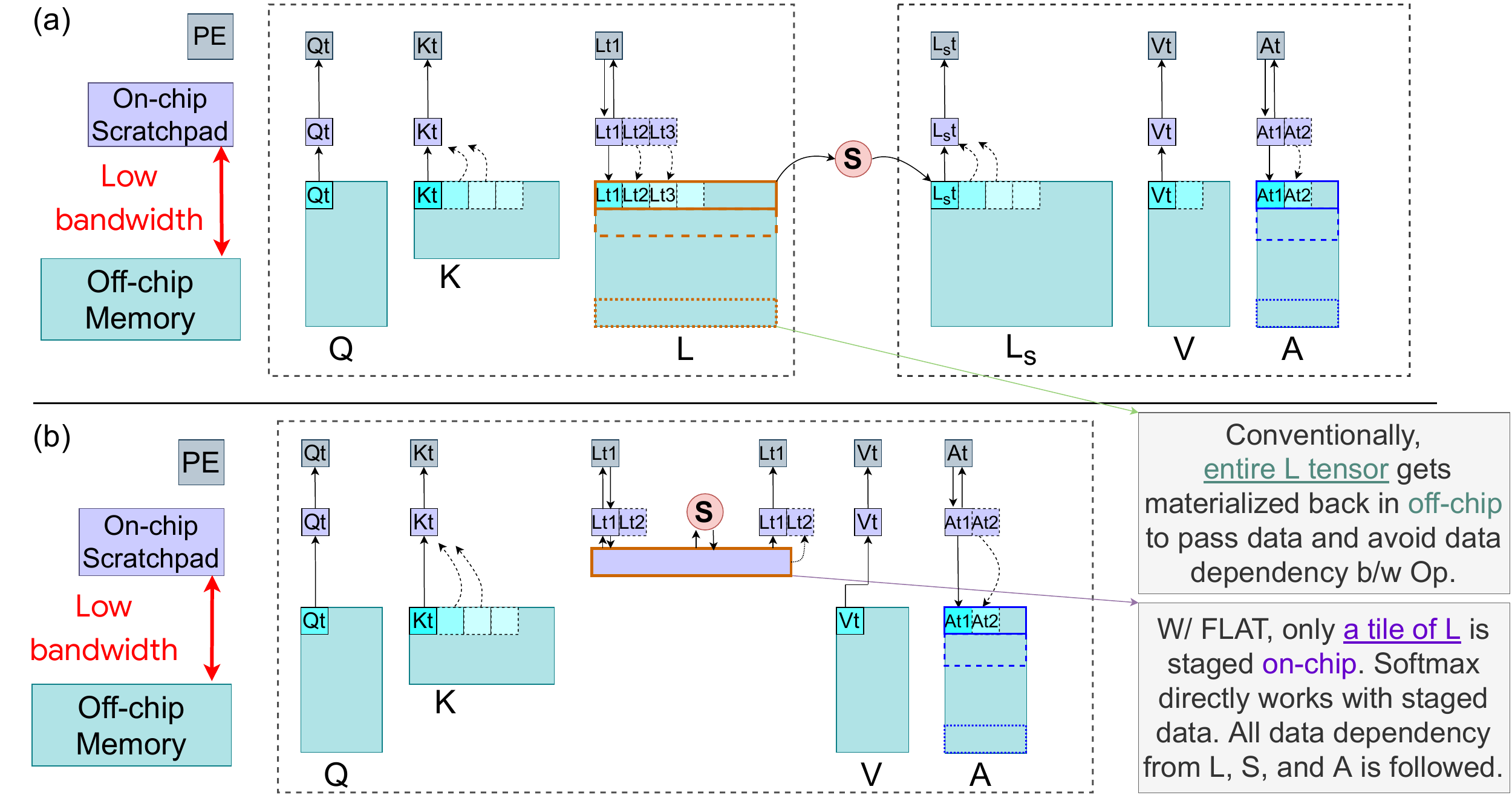}
\end{center}
\caption{(a) Baseline and (b) \dataflow dataflow. \dataflow performs inter-operator fusion of \lshort, \ashort while respecting data dependencies introduced by Softmax. \dataflow-tile enables staging slices of the logits tensor in the on-chip scratchpad increasing effective memory bandwidth. This fused, interleaved execution of \lshort, \ashort yields higher compute utilization and improved performance.}

\label{fig:flat_dataflow}
\end{figure*}

\niparagraph{Challenge 2: Effectively handling large intermediate tensors that do not fit in on-chip memory}.
Recall that the intermediate tensor between \lshort and \ashort has size $\mathcal{O}(BHN^2)$. As discussed in \autoref{sec:challenge_memfootprint}, this can easily exceed the on-chip memory capacity of DNN accelerators. Further, the specific 
size of the on-chip memory may be highly variable across different accelerators.
Owing to the above challenges, conventionally, we often do not apply tensor fusion to attention layer and stick to operator-by-operator operation scheme, as shown in \autoref{fig:flat_dataflow}(a). In this work, we use \dataflow to enable \lshort-\ashort fusion operation scheme, as in \autoref{fig:flat_dataflow}(b).
\begin{figure}[t]
\begin{center}
\includegraphics[width=0.99\linewidth]{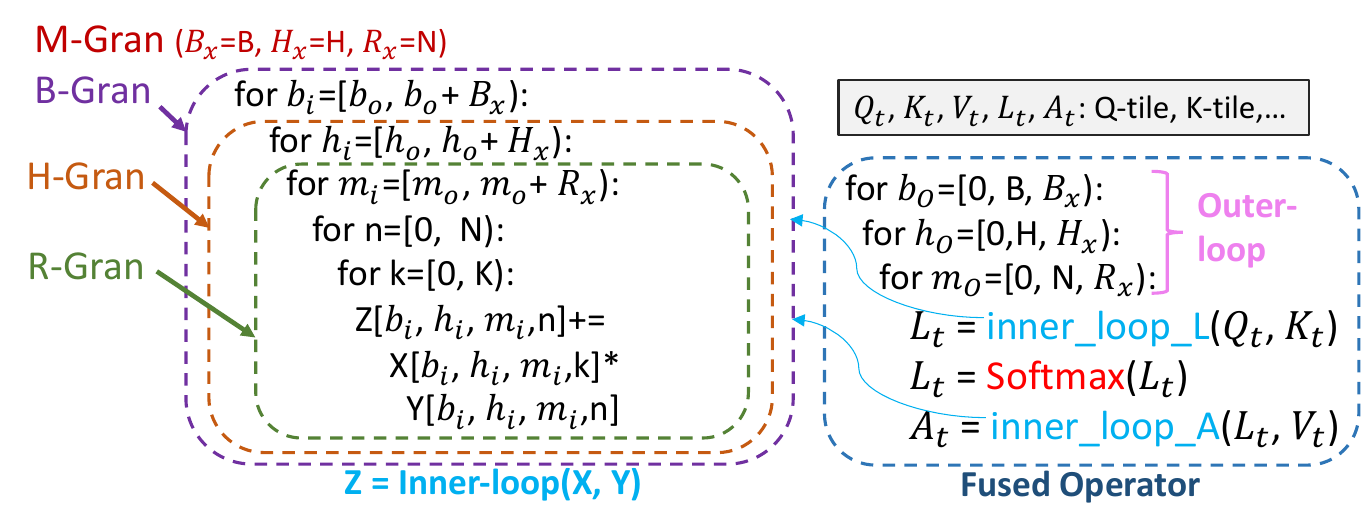}
\end{center}
\caption{For loop of fused \lshort-Softmax-\ashort (or shortened as \lshort-\ashort in the paper) and the choice of granularity.}
\label{fig:for_loop}
\end{figure}

\section{\dataflow Dataflow Implementation}
\label{sec:dataflow_implementation}
To fuse two tensor operators $X$ and $Y$, we divide the loop nests into two groups: ``\emph{outer-loop}'' and ``\emph{inner-loop}''.
We use $L$ and $A$ for illustration as shown in ~\autoref{fig:for_loop}, but the principles are applicable to any set of consecutive tensor operators. The outer-loops are shared across \lshort and \ashort. The inner-loops are unique for each operator.
After fusion, 
the fused operator has two inner-loops, which we run one after another (interleaved), and iterate through the shared outer-loop. 
Considerations for tile sizes to address the data dependence and on-chip memory constraints (\autoref{sec:fusion_challenges}) are discussed in this section.

\subsection{\dataflow-tile and Execution Granularity}
\label{subsec:granularity}
\dataflow employs two levels of tiling: intra-operator tiling and inter-operator tiling. We name each tile in inter-operator tiling, a \dataflow-tile. \dataflow computes 
\dataflow-tile activations from \lshort
and feeds it through Softmax and to \ashort.
\dataflow-tiles, the inner-loop in \autoref{fig:for_loop}, essentially specifies how many slices of the partial intermediate tensor are calculated in one pass of the fused-operator in \autoref{fig:flat_dataflow}(b).
The minimum granularity of the \dataflow-tile is determined by the data dependence constraint of Softmax and called \emph{row-granularity} (discussed in \autoref{sec:constraint_dependency}), for effectively collecting a group/tile of (input) data that fulfills the Many-to-Many dependency pattern of Softmax.
We progressively build larger (coarser-grain) tiles, namely,
tiling multiple number of rows at a time ($R_{x}$), multiple number of heads ($H_{x}$), and finally, multiple number of (micro-)batches ($B_{x}$) in the tile. We refer to these as Row (R-Gran), Head (H-Gran), and Batch (B-Gran) granularity respectively (discussed in \autoref{sec:constraint_memory}). Further, the most intuitive baseline of moving the entire intermediate tensor (namely the entire output of \lshort) on-chip is referred to as Batch-Multi-Head granularity (M-Gran), as shown in \autoref{fig:for_loop}.

\subsection{Managing Constraints from Data Dependency}
\label{sec:constraint_dependency}

\niparagraph{Basic execution unit $\rightarrow$ ``Row-granularity''.} %
The Softmax reduction is along the key dimension: this effectively captures the relative weight of each token in the query sequence against other tokens in the key sequence.
The minimum Softmax execution requires an [1, N] input array, which in turn requires a query of [1, D] and a key of [D, N], as illustrated in \autoref{fig:compute_flow} Step-\circled{2} and Step-\circled{3}\footnote{Note that here we are describing fused L-A operators, where the K, Q, and V tensors are already calculated and prepared in Step-\circled{1}.}.
This forms our basic tiling unit (finest granularity)---\textit{row-granularity}, which respects the data dependency introduced by the Softmax while keeping minimum number of elements to pass between \lshort and \ashort.
\dataflow restricts the tile sizes to operate in multiples of this row-granularity.

\subsection{Managing Constraints from On-Chip Memory}
\label{sec:constraint_memory}

\niparagraph{M-Gran, B-Gran, H-Gran: Leveraging Limited Reuse of f(\lshort, \ashort).}
\label{sec:gran_tradeoff}
Coarser granularities require staging larger tiles in the on-chip memory. As sequence lengths increase this can increase rapidly (recall the O($N^2$) growth). To fit into the limited on-chip memory, one may target finer granularities, e.g., moving from M-Gran to B-Gran (i.e., effectively tiling  micro-batches). In general, while this helps reduce the size of the tile, when we are tiling two operators at finer granularity at the outer-loop, we may trade-off the reuse opportunity at the inner-loops. For example, for f(\FC, \FC) and f(\CONV, \CONV), when decreasing the batch size (i.e., micro-batching), we directly reduce the number of times a weight can be reused. The weights need to be re-fetched again and again for each micro-batch. This effect is exacerbated when considering finer granularities such as H-Gran for the weight-activation \kshort/\qshort/\vshort/\oshort operators. The reduced reuse opportunity by inter-operator tiling reduces the achievable performance, even though the fused operator has large operational intensity (\autoref{fig:fused_quant}).    
In contrast, \lshort and \ashort are activation-activation operations (\autoref{sec:L/A operators}).
Each new activation of \lshort needs to compute with a new activation of \ashort, i.e., there are no reuse opportunities at the algorithmic level.
Decreasing the tiling granularity (M-Gran to B-Gran to H-Gran), does not preclude any reuse opportunity, since there are no reuse opportunities at the algorithmic level.
Thus, the finer M-Gran, B-Gran, H-Gran  in \dataflow are well-suited for f(\lshort, \ashort). 

\niparagraph{R-Gran: Extreme Large Sequence Range.}
\label{sec:R-gran_tradeoff}
To enable very long sequence lengths~\cite{kitaev2020reformer, compressive_transformer,performer}, but with limited on-chip memory resources~\cite{tpuv3, tesla2018v100}, we need to tile at even finer granularity, namely R-Gran.
However, finer granularities come with an associated trade-off: when we reduce the number of rows ($R_x$), we will also reduce the reuse opportunity in the matrix multiplication itself.
For example, even for \lshort/\ashort fusion, using fewer rows means the same key vectors need to be fetched multiple times across the interleaved cross-operator outer loops.
Further, reducing number of rows at the outer-loop could also decrease the achievable performance at the inner-loop, e.g., not enough dimension size to fully utilize PE array.
Thus, \dataflow co-explores inter-operator (optimizing the outer-loop) and intra-operator dataflow (optimizing the inner-loop) to mitigate these potential sources of inefficiencies. 

\niparagraph{On-chip buffer requirement.}
\autoref{table:l3_size_of_example} lists the required on-chip buffer size using \dataflow. We derive the R-Gran value here (others follow similar reasoning).
\lshort operator consumes (Rd+Nd)x2 size of the on-chip buffer (2 to account for double buffering), and \ashort consumes (Nd+Rd)x2. RN for buffering the intermediate tensor (\dataflow-tile) (no double buffering since it does not interact with off-chip memory), whose on-chip buffer requirement is shown in \autoref{table:l3_size_of_example}.

\niparagraph{HW support to implement \dataflow.}
\label{sec:hw_support}
\dataflow requires minimal HW support: (1) controller to recognize the proposed fine-grained dataflow and (2) on-chip buffer to be \textit{software-addressable} to support tiling. These features are supported by most recent industry and academic accelerators~\cite{tpuv3, simba, aimt, tesla2018v100}.
\begin{table}[t!]
\footnotesize
\centering
\vspace*{0.1cm}
\caption{\label{table:l3_size_of_example}Buffer requirement for tiling granularity. M: batched Multi-head, B: Batch, H: Head, R: Row.}
\resizebox{0.48\textwidth}{!}{
\begin{tabular}{l|c|c|c|c}
\toprule
\textbf{Granularity} & \textbf{M-Gran} & \textbf{B-Gran} & \textbf{H-Gran} & \textbf{R-Gran}\\\bottomrule
Buffer Req. & $\mathcal{O}(8BDN+BHN^{2})$ & $\mathcal{O}(8DN+HN^{2})$ & $\mathcal{O}(8Nd+N^{2})$ & $\mathcal{O}(4Rd+4Nd+RN)$\\\bottomrule
\end{tabular}
}
\end{table}
\section{Evaluation Methodology}
\label{sec:analytical_model}
\begin{figure}
\begin{center}
\includegraphics[width=0.99\linewidth]{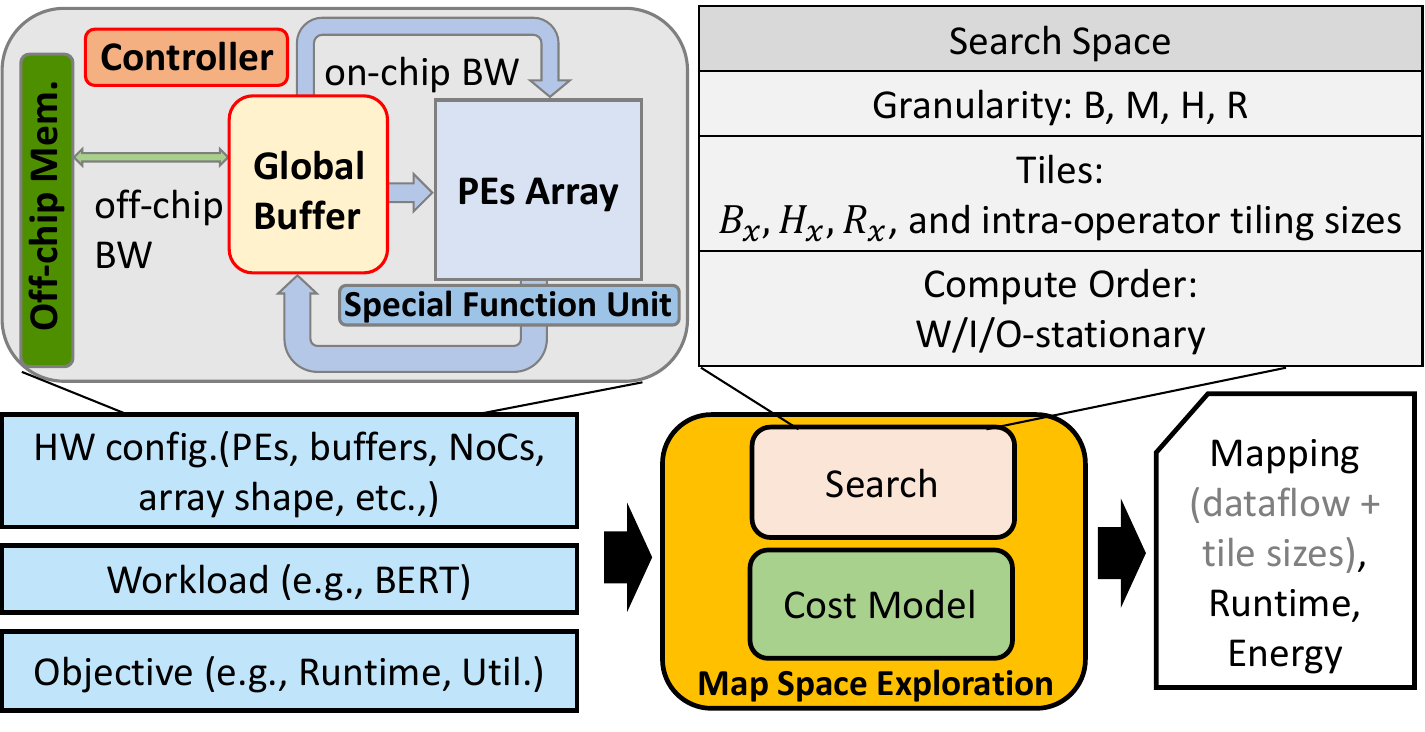}
\end{center}
\vspace{-0.4cm}
\caption{Map space exploration framework. (Special Function Unit: for computing non-linear operations, e.g., softmax, activation.)}
\label{fig:analytical_model}
\end{figure}
\subsection{Accelerator Modeling Methodology}
\label{sec:accelerator_models}
We developed a detailed analytical cost model to estimate the performance and energy consumption of \dataflow across a range of hardware accelerators configurations,  following similar methodology as prior work~\cite{timeloop,maestro}. 
We meticulously model the major microarchitectural blocks commonly shared by most DNN accelerators as outlined in Figure~\ref{fig:analytical_model}.
Based on this model, we collect the relevant architectural details, which later are used to compute the accelerator performance metrics.

\niparagraph{Compute model.}
We model the compute array as a collection of processing elements with configurable bandwidth from/to the global on-chip buffers.
The compute array model supports common intra-operator dataflow, including weight, input, and output stationary.
In addition, we model various data distribution and reduction NoCs, including systolic, tree, or crossbar structures to study the the trade-offs between compute bandwidth and distribution-collection time~\cite{maestro, maeri}. 
Following this methodology, we model TPU~\cite{tpu} (systolic-array) as well as other spatial array accceleraors, such as Eyeriss v2~\cite{eyeriss_v2} and MAERI~\cite{maeri}).
We also carefully model the overhead of switching tiles for filling and draining data to reflect the cold start and tailing effect. Finally, we account for softmax operation runtime in all the evaluations.

\niparagraph{Buffering model.}
Studying dataflow optimization techniques demand for a detailed modeling of buffers.
To achieve this objective, we model PE arrays with local scratchpad for input, weight, intermediate results, and output storage.
We add the on-chip global buffer to store the intra- and inter-operator tiles.
The performance model also includes the data spilling overhead. That is when the live memory footprint (buffer requirement for staging data on-chip) is larger than the on-chip global buffer capacity.

\niparagraph{Memory bandwidth.}
Since there are multiple microarchitectural units that access the on-chip and off-chip memories, we model them as limited bandwidth shared-hardware resources.
That is, if the access rate to a shared memory resource exceeds a pre-defined bandwidth, the data accesses are throttled. This overhead manifests as longer runtime.
A key feature of our simulation methodology is the detailed modeling of the accelerator memory hierarchy to systematically assess the memory-boundedness of attention operators and their pressure on off-chip memory bandwidth.

\niparagraph{Energy model.}
Collecting the detailed activity counts from the analytical model, we use Accelergy~\cite{accelergy} framework to estimate the energy consumption for the major microarchitectural blocks. That includes compute, on-chip memory, and off-chip memory communications.
Note that \dataflow neither alters the total number of computations nor the total number of accesses to the on-chip global buffer. Instead, it optimizes the number of off-chip memory accesses, which is the major contributor to the overall accelerator energy consumption~\cite{synthesislectures_efficient, chen2016eyeriss}.

\niparagraph{Map-space exploration workflow.}
\label{sec:dataflow_dse}
We also integrate a map-space exploration (MSE) workflow (\autoref{fig:analytical_model}) into our simulation framework. 
The main purpose of this exploration workflow is to carry out a search algorithm in a predefined map space governed by the cost model.
In this work, we use exhaustive search to find the optimal design point uniformly across all the dataflow optimizations.
MSE includes both intra- and inter-operator dataflow optimization space (enabling optimal dataflow comparisons with and without \dataflow technique later in \autoref{table:df_config} and \autoref{sec:evaluations}). The relevant architectural parameters for this optimization space are outlined in \autoref{fig:analytical_model}.

\niparagraph{Comparison to prior accelerator modeling tools.}
There are several popular open-sourced DNN accelerator modeling frameworks~\cite{timeloop, maestro, yang2020interstellar,dmazerunner,scalesim,zigzag}.
However, none of them offer support for cross-layer performance (and reuse) modeling, assuming layer-by-layer execution.
In contrast, our framework evaluates the performance of DNN models in both single-layer and cross-layer manner, enabling various cross-operator fusion studies.
To ensure the integrity and correctness of our framework, we compared the simulation results from our framework under single-layer modeling to MAESTRO~\cite{maestro}. The performance metrics are within $1\%$ difference to MAESTRO's results.

\niparagraph{Target accelerator configurations.}
We evaluate the benefits of \dataflow on two different accelerator regimes, namely cloud and edge accelerators.
As outlined in \autoref{table:df_accel_config}, we set the accelerator configurations in our model following the designs proposed for cloud~\cite{tpuv3,tesla2018v100} and edge~\cite{eyeriss_v2,yazdanbakhsh2021evaluation,edgetpu} accelerators.
In all the evaluations, we allot sufficient FLOPs to the Special Function Unit (\autoref{fig:analytical_model}) in order to eradicate the expected compute bottlenecks, uniformly across all the dataflow variants.

\niparagraph{Evaluation metrics.}
\label{sec:util_rate}
For all the evaluations, we use performance and energy savings as efficiency metrics.
For comparisons between different models, we normalize the runtime of each dataflow by the ideal runtime of the target workload as follows:
\vspace{-0.2cm}
$$ Util = \frac{\mathrm{Runtime}_{ideal}}{\mathrm{Runtime}_{dataflow}}$$
\noindent ; where $\mathrm{Runtime}_{ideal}$ is the arithmetic optimal runtime of the current workload. That is, the total computes in a model divided by the peak FLOPs of the target accelerator. $\mathrm{Runtime}_{dataflow}$ represents the achieved runtime by a dataflow optimization.
This \textit{normalized runtime} metric explains how far the current dataflow is from its arithmetic optimum. This metric is an indication of the distance to the dataflow compute-boundray in the roofline model as well as compute resource utilization ($Util$).

\niparagraph{\dataflow on GPUs.}
\label{sec:flat_gpu}
Some of our evaluations demonstrate the efficiency of FLAT on Nvidia-Tesla-T4~\cite{teslaT4} GPU with 16GB memory.
Since we could not modify the underlying highly-optimized CUDA APIs, we manually implemented the ``einsum'' operation as nested loops for baseline. We prototyped fused L-A by modifying this nested loop.

\niparagraph{Workloads.}
We study a range of recent attention-based models, including BERT-Base~\cite{transformer} (\bench{BERT}), TransformerXL~\cite{TrXL} (\bench{TrXL}), FlauBERT~\cite{le2019flaubert} (\bench{FlauBERT}), T5~\cite{t5_model} (\bench{T5}), and XLM-MLM-En~\cite{xlm_model} (\bench{XLM}).
We evaluate these models under different sequences lengths ranging from $N=512$ to $N=64K$ to imitate attention-based models with long sequence length~\cite{beltagy2020longformer, kitaev2020reformer}.
We also study a future-proofing sequence length of size $N=256K$.
We use a batch size of 64 for all the models. Note that the batch size choice is immaterial to our dataflow optimization.
\begin{table}[t!]
\footnotesize
\centering
\caption{\label{table:df_accel_config}The HW compute resource and BW configuration of Edge and Cloud platforms in the evaluation sections. The on-chip memory capacity is varied across explored design-points.}
\resizebox{0.48\textwidth}{!}{
\renewcommand{\aboverulesep}{0pt}
\renewcommand{\belowrulesep}{0pt}
\begin{tabular}{l|c|c|c}
\toprule
\textbf{Platform} & \textbf{$\#$ of PEs} & \textbf{On-Chip BW} & \textbf{Off-Chip BW}\\\bottomrule
Edge & 32$\times$32&1 TB/Sec&50 GB/Sec\\\hline
Cloud & 256$\times$256&8 TB/Sec&400 GB/Sec\\\bottomrule
\end{tabular}
}
\end{table}
\begin{table}[t!]
\footnotesize
\centering
\caption{\label{table:df_config}Comparisons dataflow configurations.}
\resizebox{0.48\textwidth}{!}{
\renewcommand{\aboverulesep}{0pt}
\renewcommand{\belowrulesep}{0pt}
\begin{tabular}{p{0.16\textwidth}|c|p{0.3\textwidth}}
\toprule
\textbf{Dataflow} & \textbf{Design Point} & \textbf{Description}\\\bottomrule
\makecell[l]{Na\"ive\\(Intra-Operator)}& Na\"ive&Intra-operator weight-stationary dataflow with fixed tile size, similar to~\cite{edgetpu,du2015shidiannao}.\\\hline
\makecell[l]{\flex\\(Intra-Operator)} & \flex-Opt & We exhaustively search for optimal intra-operator dataflow, reflecting the optimal solution can be found in existing intra-operator mappers~\cite{nvdla,eyeriss_v2,gamma,tpuv3}.\\\hline
\makecell[l]{\dataflow\\(Intra-/Inter-Operator)}& \dataflow-Opt&We exhaustively search for optimal intra-operator as well as inter-operator dataflow.\\\bottomrule
\end{tabular}
}
\end{table}

\begin{table}[t]\centering
\caption{Run time performance improvement of \dataflow over Na\"ive and \flex, using sequence length=512 and Edge platform compute + BW configurations (\autoref{table:df_accel_config}) with varying on-chip buffer sizes (200K, 20M and 2GB). \dataflow shows its advantage when buffer sizes are limited.}
\scriptsize
\resizebox{0.48\textwidth}{!}{
\renewcommand{\aboverulesep}{0pt}
\renewcommand{\belowrulesep}{0pt}
\begin{tabular}{l|rrr|rrrr}\toprule
\multirow{2}{*}{\makecell{Run time \\ Improvement}} &\multicolumn{3}{c|}{L/A layer} &\multicolumn{3}{c}{End-to-End} \\\cmidrule{2-7}
&2G(B) &20M &200K &2G&20M &200K \\\bottomrule
\dataflow over Na\"ive &1.7$\times$ &3.3$\times$ &3.2$\times$ &1.5$\times$ &1.6$\times$ &3.2$\times$ \\\midrule
\dataflow over \flex &1.02$\times$ &1.02$\times$ &1.7$\times$ &1.02$\times$ &1.02$\times$ &1.1$\times$ \\
\bottomrule
\end{tabular}
}
\label{table:exp_summary}
\end{table}
\begin{figure}[t]
    \centering
    \subfloat[Sequence Length = 512]{
    \includegraphics[width=0.49\textwidth]{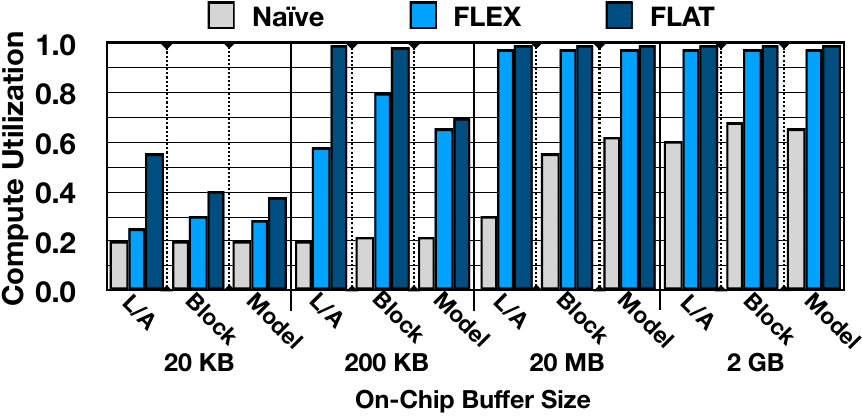}
    \label{fig:util_cpr_edge_a}}\\
    \subfloat[Sequence Length = 4\,K]{
    \includegraphics[width=0.49\textwidth]{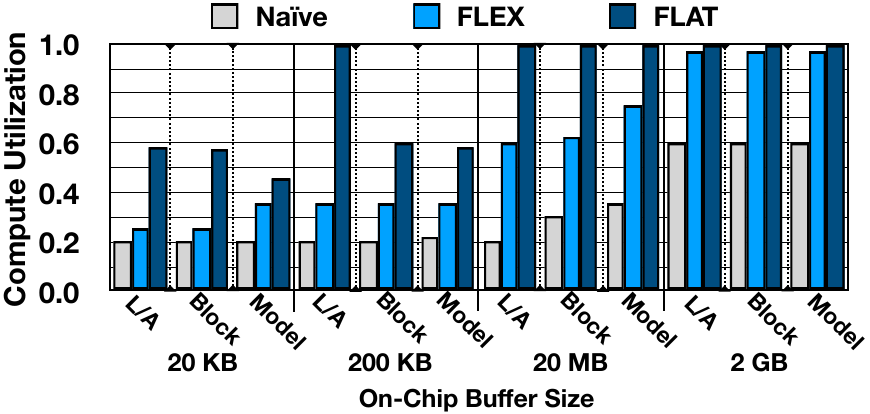}
    \label{fig:util_cpr_edge_b}}\\
    \subfloat[Sequence Length = 64\,K]{
    \includegraphics[width=0.49\textwidth]{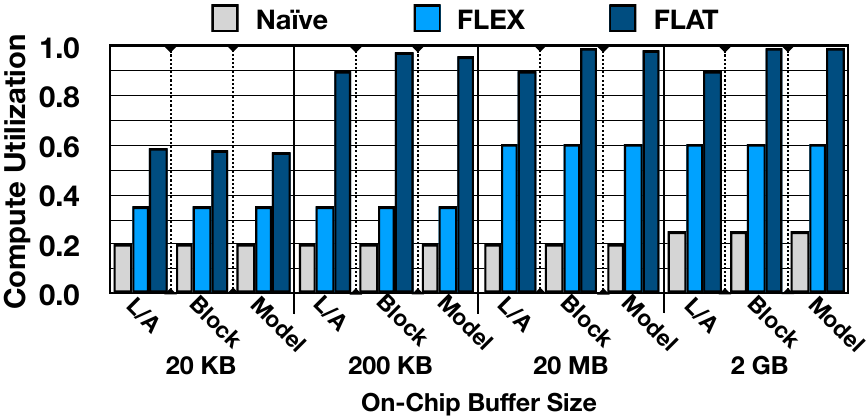}
    \label{fig:util_cpr_edge_c}}
    \caption{Comparisons of compute utilization of BERT under Edge platform w/ different input sequence lengths. We sweep the size of available on-chip buffer from 20KB to 2GB. The figures demonstrate three different performance analysis, (first bar) L-A $\rightarrow$ focusing on performance difference at the L, A operators; (second bar) Block $\rightarrow$ consider all operators in the attention block; and (third bar) Model $\rightarrow$ a model-wise (end-to-end) performance.}
    \label{fig:util_cpr_compute_edge}
\end{figure}

\begin{figure}[t]
    \centering
    \subfloat[Sequence Length = 512]{
    \includegraphics[width=0.49\textwidth]{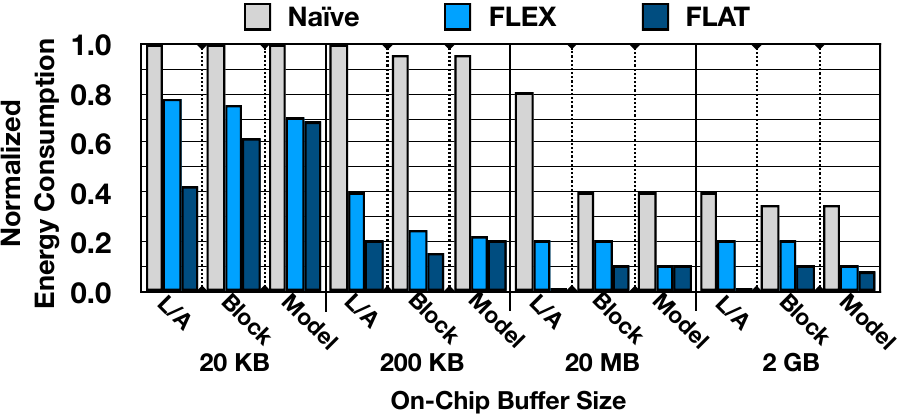}
    \label{fig:util_cpr_edge_d}}\\
    \subfloat[Sequence Length = 4\,K]{
    \includegraphics[width=0.49\textwidth]{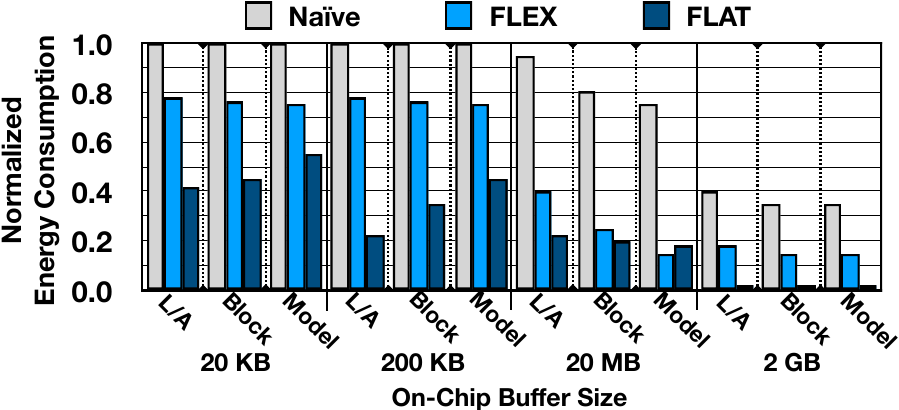}
    \label{fig:util_cpr_edge_e}}\\
    \subfloat[Sequence Length = 64\,K]{
    \includegraphics[width=0.49\textwidth]{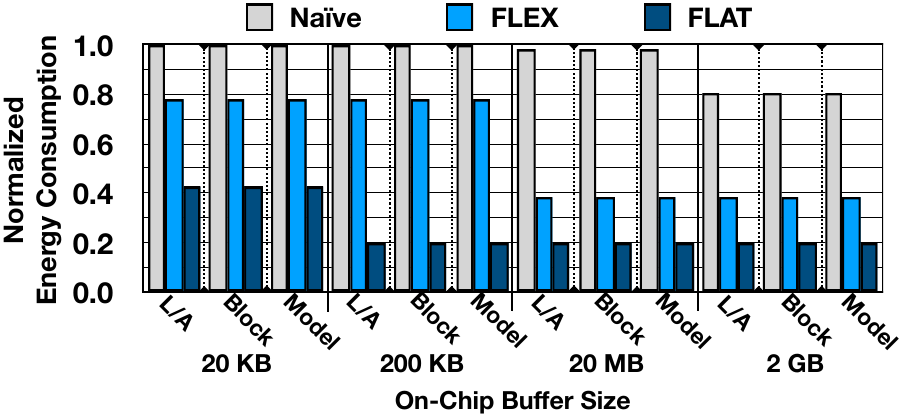}
    \label{fig:util_cpr_edge_f}}
    \caption{Comparisons of energy consumption of BERT under Edge platform with different input sequence length. We normalize the energy consumption results to the largest value in each sub-plot. Each bar represents the same analysis as described in Figure~\ref{fig:util_cpr_compute_edge}.}
    \label{fig:util_cpr_energy_edge}
\end{figure} 
\begin{figure}[t]
    \centering
    \subfloat[Sequence Length = 512]{
    \includegraphics[width=0.47\textwidth]{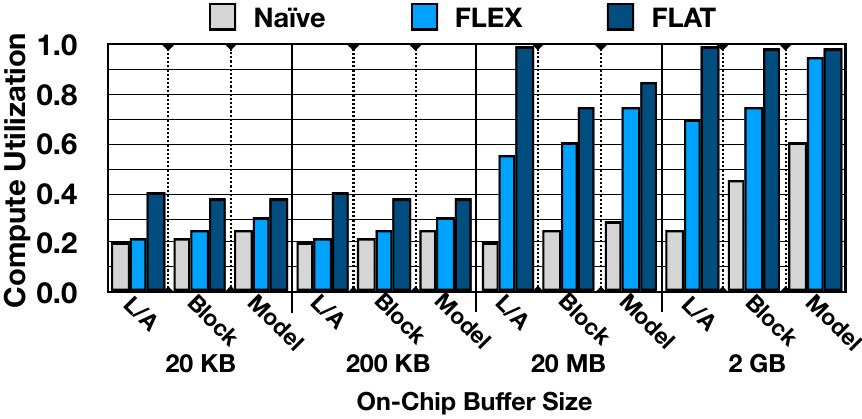}
    \label{fig:util_cpr_cloud_a}}\\
    \subfloat[Sequence Length = 4\,K]{
    \includegraphics[width=0.47\textwidth]{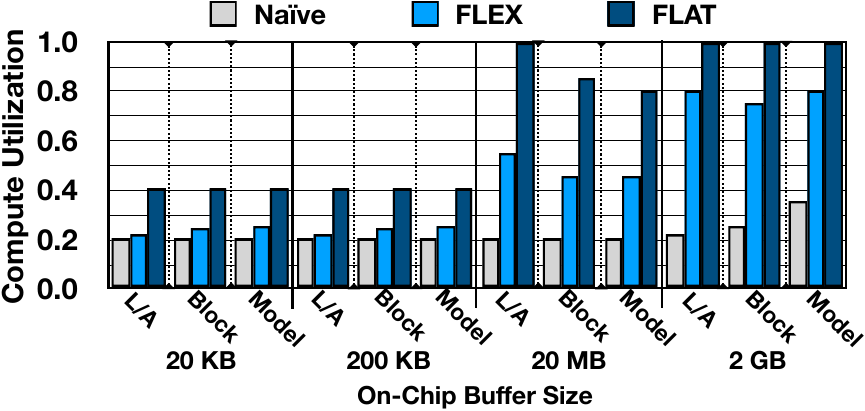}
    \label{fig:util_cpr_cloud_b}}\\
    \subfloat[Sequence Length = 64\,K]{
    \includegraphics[width=0.47\textwidth]{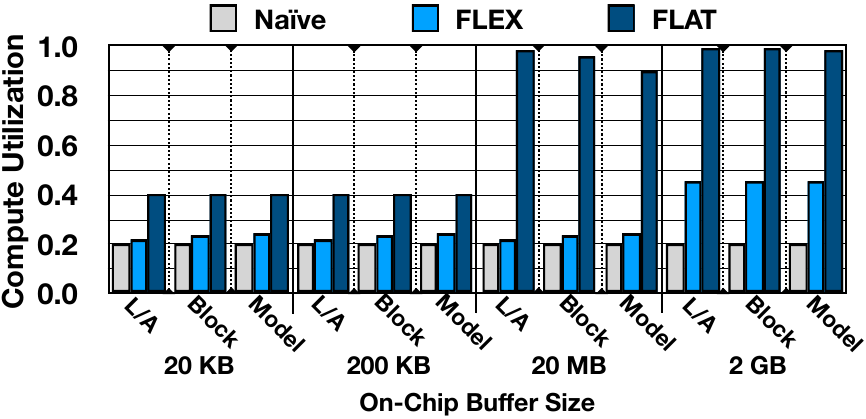}
    \label{fig:util_cpr_cloud_c}}
    \caption{Comparisons of compute utilization of XLM under Cloud platform with different input sequence length. We sweep the size of available on-chip buffer from 20KB to 2GB. Each bar represents the same analysis as described in Figure~\ref{fig:util_cpr_compute_edge}.}
    \label{fig:util_cpr_compute_cloud}
\end{figure} 
\begin{figure}[t]
    \centering
    \subfloat[Sequence Length = 512]{
    \includegraphics[width=0.49\textwidth]{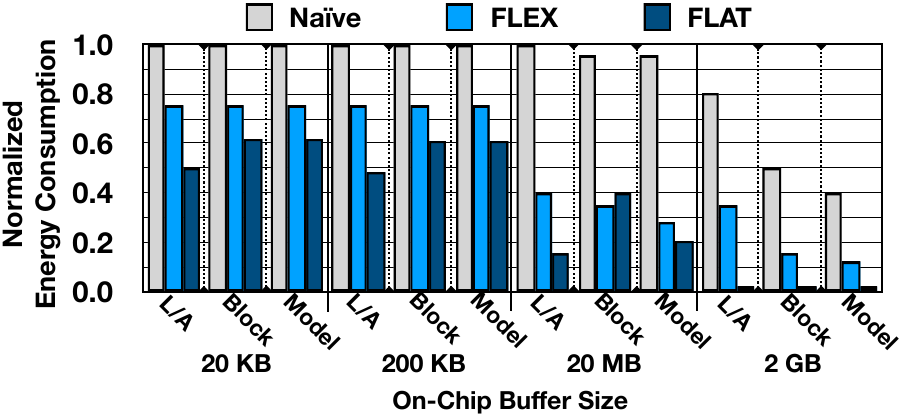}
    \label{fig:util_cpr_cloud_d}}\\
    \subfloat[Sequence Length = 4\,K]{
    \includegraphics[width=0.49\textwidth]{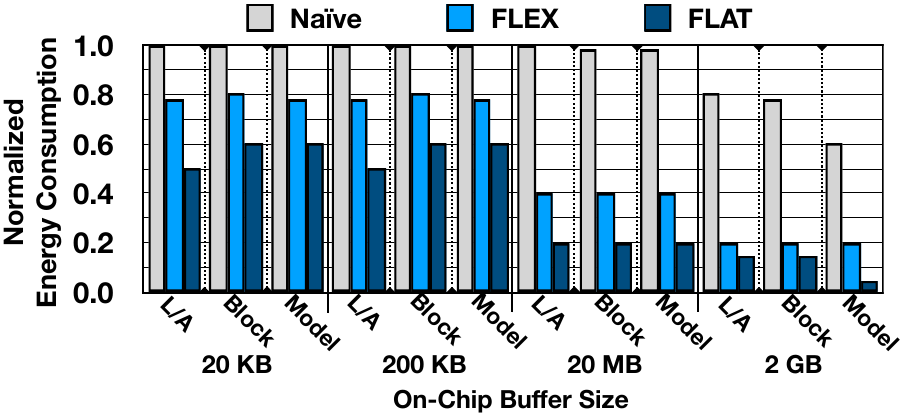}
    \label{fig:util_cpr_cloud_e}}\\
    \subfloat[Sequence Length = 64\,K]{
    \includegraphics[width=0.49\textwidth]{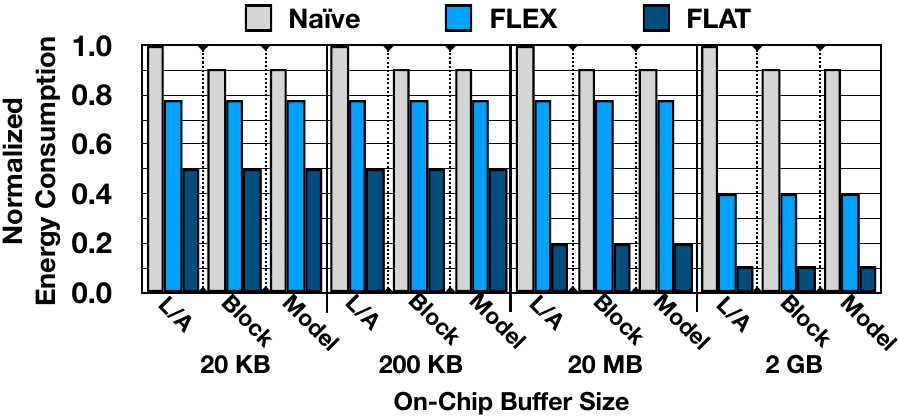}
    \label{fig:util_cpr_cloud_f}}
     \caption{Comparisons of energy consumption of XLM under Cloud platform with different input sequence length. We normalize the energy consumption results to the largest value in each sub-plot. Each bar represents the same analysis as described in Figure~\ref{fig:util_cpr_compute_edge}.}
    \label{fig:util_cpr_energy_cloud}
\end{figure} 

\section{Evaluation I: \dataflow Dataflow Efficacy}
\label{sec:evaluations}
\rev{In this section, we fix the “headline” ``HW resources” (i.e., FLOPs and off-chip memory bandwidth) as outlined in \autoref{table:df_accel_config} and sweep-and-explore other microarchitectural parameters relevant to the dataflow efficiency, including on-chip memory size and dataflow variations (\naive, \flex and \dataflow as explained in \autoref{table:df_config}).
The on-chip memory size assesses the dataflow optimizations associated to the large intermediate tensor size in the attention layers.
The goal of this section is demonstrate the benefits of \dataflow across a range of hardware and dataflow configurations, without biasing to any specific design point.}

\niparagraph{Runtime performance.}
\label{sec:exp_util}
\autoref{table:exp_summary} shows the run time performance improvement of \dataflow (\dataflow-Opt) over \naive (\textit{the baseline dataflow without any optimizations}) and \flex (\flex-Opt), under commonly observed on-chip buffer sizes and sequence length (512-token).
We observe 1) providing huge on-chip buffer resources (2GB), \dataflow with its fused operation and improved data-reuse can improves \naive and \flex; 2) at limited buffer resources, \dataflow becomes handy owing to its reduced on-chip buffer footprint. For example, at 200KB, \dataflow-Opt improves the current state-of-the-art baseline -- \flex-Opt -- by 1.7$\times$ on the focusing L-A operations, which contributes to 1.1$\times$ improvement of end-to-end performance.
Note that owing to the quadratic complexity of L-A operations, L-A operations will become more dominant at larger sequence length.
For e.g., in this evaluation with 512-token, L-A operation contributes only 10$\%$ of the computes to the end-to-end model; however, it increases to 47$\%$ and 78$\%$ at 4K-token and 16K-token.
Next, we show the efficacy of \dataflow under a sweep of buffer sizes, sequence length, and with perspectives of different granularity of the models (operations, layers, and end-to-end).
%


\niparagraph{Sensitivity to sequence length.}
As described in \autoref{sec:util_rate}, we use $Util$, a normalized run time performance metric to show the performance difference of different sequence lengths, buffer size, and model granularity in the same plot.
As \autoref{fig:util_cpr_edge_a}-\subref*{fig:util_cpr_edge_c} shows, \dataflow-Opt consistently outperforms \naive and \flex-Opt.
Analyzing the results indicate that though tensor-tensor fusion seems to be complicated and deemed as non-profitable, \dataflow can efficiently execute tensor-tensor fusion in attention layers and harvest the highest performance gains.
In \autoref{fig:util_cpr_edge_a}-\subref*{fig:util_cpr_edge_c}, as the sequence length increases, the on-chip buffer requirement increases quickly (\autoref{table:l3_size_of_example}).
Under this scenario, most of the accelerator design points in \flex design space starts to hit the memory boundedness.
However, applying the \dataflow technique, we can effectively reduce the memory requirement and thus providing a better scalability to sequence length.
At the optimal design point (\dataflow-Opt) reaches nearly 1.0 compute utilization with 10$\times$-100$\times$ less on-chip buffer requirement \rev{in Edge accelerator as shown in \autoref{fig:util_cpr_edge_a}-\subref*{fig:util_cpr_edge_c}, and  100$\times$-1000$\times$ in Cloud accelerator as shown in \autoref{fig:util_cpr_cloud_a}-\subref*{fig:util_cpr_cloud_c}, a scarce and critical hardware resource for accelerators, compared to \flex-Opt.}



\begin{figure}[ht]
\begin{center}
\includegraphics[width=0.97\linewidth]{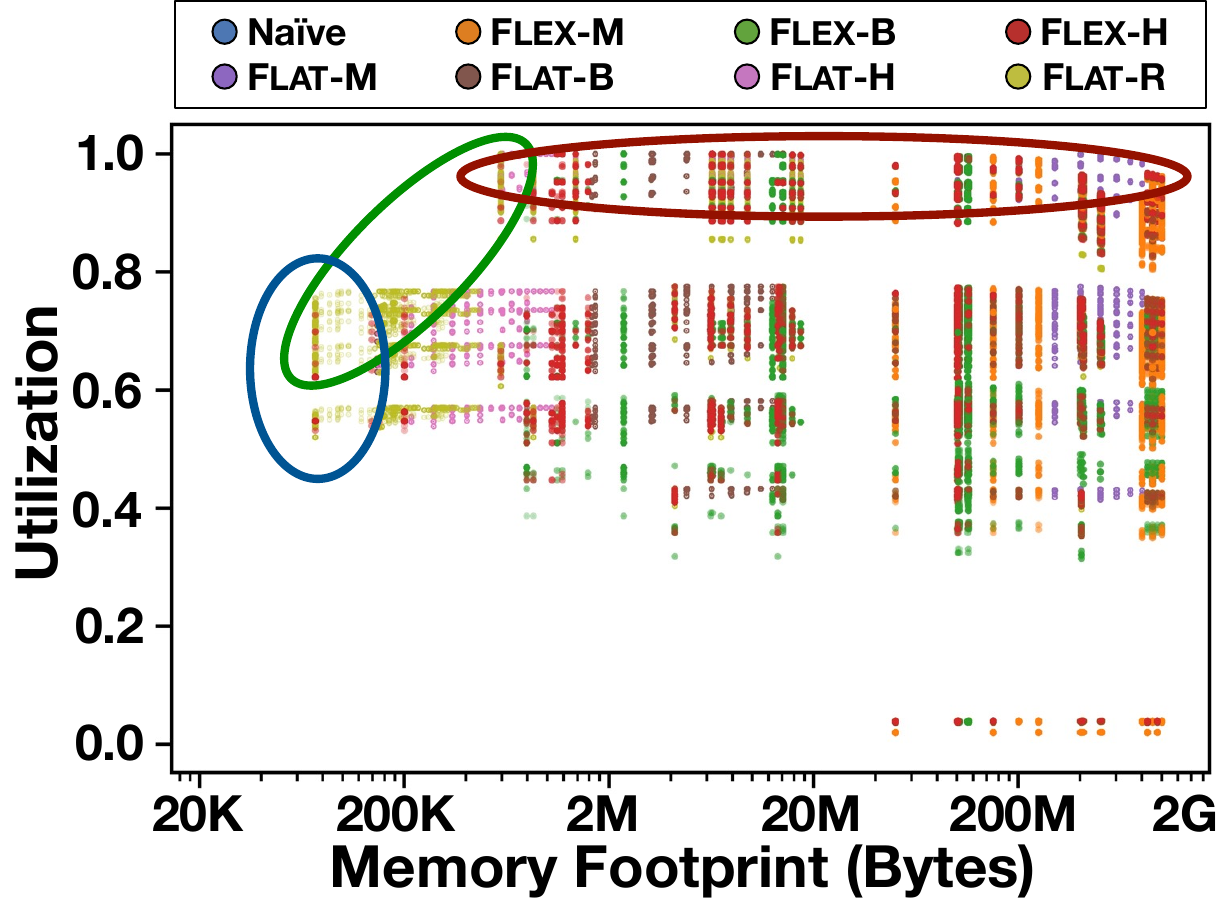}
\vspace{-0.5cm}
\end{center}
\caption{The design space of \dataflow of BERT w/ input sequence length of 512 in Edge. \flex dataflow (\flex-X): \flex dataflow with X-granularity. \dataflow dataflow (\dataflow-X): \flex dataflow with X-granularity, where X could be M (batch-Multi-head), B (Batch), H (head), or R (row). The design-point with the highest utilization, given a buffer constraint represents \flex-Opt and \dataflow-Opt (\autoref{table:df_config}). The highlighted regions represent: ``{\setlength{\fboxrule}{0.75pt}\setlength{\fboxsep}{0.2pt}\fcolorbox{plotblue}{white}{Blue (bottom left)}}'' $\rightarrow$ low memory footprint, ``{\setlength{\fboxrule}{0.75pt}\setlength{\fboxsep}{0.2pt}\fcolorbox{plotgreen}{white}{Green (top left)}}'' $\rightarrow$ high utilization per memory footprint, and ``{\setlength{\fboxrule}{0.75pt}\setlength{\fboxsep}{0.2pt}\fcolorbox{plotred}{white}{Red (top)}}'' $\rightarrow$ high utilization.}
\label{fig:dse_edge_bert_512}
\end{figure}

\begin{figure}
    \centering
    \subfloat[BERT on Edge Accelerator]{
    \includegraphics[width=0.49\textwidth]{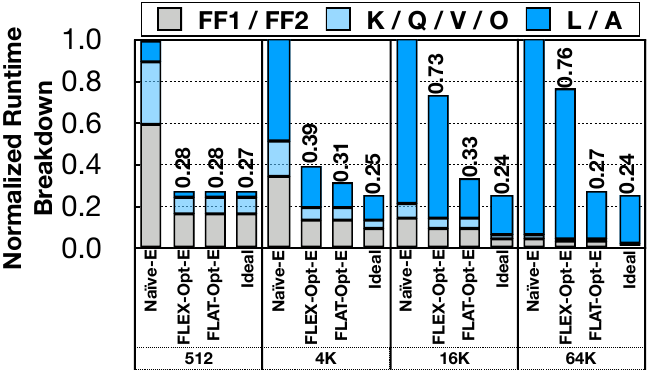}
    \label{fig:latency_ratio_a}}\\
    \subfloat[XLM on Cloud Accelerator]{
    \includegraphics[width=0.49\textwidth]{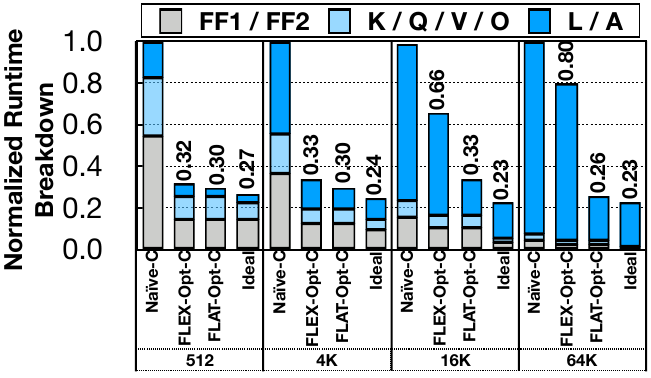}
    \label{fig:latency_ratio_b}}
    \caption{End-to-end latency breakdown. The suffix ``E'' and ``C'' indicate E(dge) and C(loud) platforms. Na\"ive, FLEX-Opt, and FLAT-Opt are defined in \autoref{table:df_config}.}
    \label{fig:latency_ratio}
\end{figure}
\begin{table*}[!t]
\centering
\caption{The End-to-End speedup and energy-consumption ratio of \dataflow-Opt-E(dge) over \flex-Opt-E(dge) and \dataflow-Opt-C(loud) over \flex-Opt-C(loud) on different models.}
\renewcommand{\arraystretch}{1.0}
\tiny{
\resizebox{0.99\textwidth}{!}{
\setlength\tabcolsep{5pt}
\renewcommand{\aboverulesep}{0pt}
\renewcommand{\belowrulesep}{0pt}
\begin{tabular}{l|c|c|c|c|c||c|c|c|c|c}\toprule[0.6pt]
\multicolumn{11}{c}{\textbf{\dataflow-Opt-Edge vs. \flex-Opt-Edge}}\\\toprule[0.6pt]
\textbf{Edge}&\multicolumn{5}{c||}{Geomean Speadup = \textbf{1.75}$\times$}&\multicolumn{5}{c}{Geomean Consumption Ratio = \textbf{0.56}$\times$}\\\midrule
\textbf{Seq. Length}&512&4K&16K&64K&256K&512&4K&16K&64K&256K\\\toprule[0.6pt]
\benchtiny{BERT}&1.02&1.27&2.21&2.84&3.10&0.98&0.78&0.44&0.34&0.31\\\midrule
\benchtiny{TrXL}&1.02&1.23&2.06&2.75&3.07&0.98&0.81&0.48&0.35&0.31\\\midrule
\benchtiny{FlauBERT}&1.01&1.11&1.62&2.26&2.67&1.00&0.90&0.61&0.43&0.36\\\midrule
\benchtiny{T5}&1.03&1.34&2.40&2.93&3.13&0.97&0.74&0.41&0.33&0.31\\\midrule
\benchtiny{XLM}&1.00&1.05&1.35&1.87&2.38&1.00&0.95&0.74&0.52&0.31\\\toprule[0.6pt]
\CC{}\textbf{Average}&\CC{}\textbf{1.02}&\CC{}\textbf{1.20}&\CC{}\textbf{1.89}&\CC{}\textbf{2.50}&\CC{}\textbf{2.85}&\CC{}\textbf{0.99}&\CC{}\textbf{0.83}&\CC{}\textbf{0.52}&\CC{}\textbf{0.39}&\CC{}\textbf{0.32}\\\toprule[0.6pt]
\multicolumn{11}{c}{\textbf{\dataflow-Opt-Cloud vs. \flex-Opt-Cloud}}\\\toprule[0.6pt]
\textbf{Cloud}&\multicolumn{5}{c||}{Geomean Speedup =  \textbf{1.65}$\times$}&\multicolumn{5}{c}{Geomean Energy Consumption Ratio = \textbf{0.45}$\times$}\\\midrule
\textbf{Seq. Length}&512&4K&16K&64K&256K&512&4K&16K&64K&256K\\\toprule[0.6pt]
\benchtiny{BERT}&1.16&1.38&1.46&2.23&2.72&0.71&0.68&0.11&0.34&0.27\\\midrule
\benchtiny{TrXL}&1.13&1.34&1.45&2.20&2.71&0.73&0.27&0.13&0.35&0.27\\\midrule
\benchtiny{FlauBERT}&1.07&1.21&1.42&2.21&2.93&0.87&0.80&0.72&0.49&0.37\\\midrule
\benchtiny{T5}&1.18&1.43&1.48&2.26&2.73&0.69&0.66&0.50&0.33&0.27\\\midrule
\benchtiny{XLM}&1.02&1.06&1.13&1.98&3.09&0.97&0.89&0.78&0.50&0.31\\\toprule[0.6pt]
\CC{}\textbf{Average}&\CC{}\textbf{1.11}&\CC{}\textbf{1.28}&\CC{}\textbf{1.38}&\CC{}\textbf{2.17}&\CC{}\textbf{2.83}&\CC{}\textbf{0.79}&\CC{}\textbf{0.61}&\CC{}\textbf{0.33}&\CC{}\textbf{0.40}&\CC{}\textbf{0.30}\\\toprule[0.6pt]
\end{tabular}
}
}
\label{fig:exp_accel}
\end{table*}

\begin{figure}
\begin{center}
\includegraphics[width=0.99\linewidth]{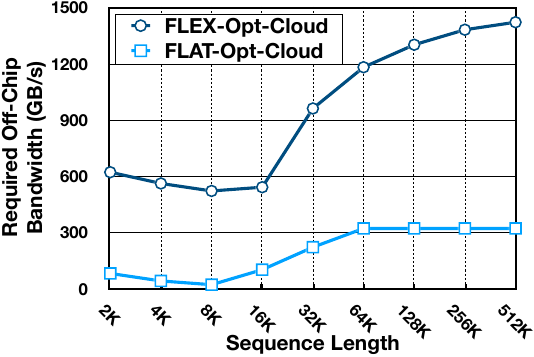}
\end{center}
\caption{The required off-chip bandwidth to reach a compute utilization rate higher than 0.95 in the most BW-intensive L-A operator when running XLM.}
\label{fig:exp_accel_fig}
\end{figure}

\niparagraph{Sensitivity to end-to-end performance.}
So far, we analyze the performance for only \lshort and \ashort operators, while 
not considering other operators within the model.
As shown in the first row of \autoref{fig:util_cpr_compute_edge} from left to right, we observe that the effect of \lshort/\ashort operators are diluted as more operators are considered.
In attention-based models, \FC/\GEMM and attention operators, namely \lshort and \ashort, are generally the most dominant computation.
For \FC/\GEMM, the typical single (intra-)operator dataflow is often sufficient to reach a high compute utilization, and hence \dataflow-Opt and \flex-Opt performs equally well for these operators.
As we can see, for the sequence length below 512, both Block-level and Model-level (i.e., End-to-End) performance is dominated by \FC/\GEMM operators. Therefore, the gains from \flex-Opt and \dataflow-Opt are immaterial.
The significant gains from our approach emerge when the sequence length increases beyond 512 to 4K, 16K, and to 64K.
Under these long-sequence lengths, the runtime contribution of \lshort and \ashort operators grows from 12$\%$ to 49$\%$, 79$\%$, and 94$\%$, respectively. This increase causes our proposed \dataflow-Opt to outperform \flex-Opt significantly even in Block and Model level scenarios.

\niparagraph{Energy consumption.}
\label{sec:exp_energy}
\autoref{fig:util_cpr_edge_d}-\subref*{fig:util_cpr_edge_f} and \autoref{fig:util_cpr_cloud_d}-\subref*{fig:util_cpr_cloud_f} show the energy consumption for \bench{BERT} on Edge platform and \bench{XLM} on Cloud platform, respectively.
It is worth to mention that high utilization does not directly translate to better energy savings; however, highly correlated.
Data points with high compute utilization generally employ better memory access patterns (e.g., less off-chip memory access and better data reuse) and thus impose less cost in terms of memory access energy, the dominant contributor to the overall energy consumption of DNN accelerators.
We observe that \dataflow-Opt reduces the energy consumption by around 1.5$\times$-2.0$\times$ comparing to \flex-Opt. 

\niparagraph{Map space exploration.}
\label{sec:exp_MSE}
\autoref{fig:dse_edge_bert_512} shows a holistic view of the entire design space of \dataflow dataflow.
The top-left corner of the diagram indicates high utilization with the least memory footprint.
For each dataflow, there are abundance of parameters that can be tuned under different optimization objectives and design constraints.
For example, while in this work, we focus on maximizing the compute utilization, one may choose other objectives such as maximizing utilization normalized to memory footprint size, leading to points in the top-left corner, or the least memory footprint size, leading to points in the left-most region.
From \autoref{fig:dse_edge_bert_512}, we can see that different dataflow configurations in the design space indeed represent notable differences in performance and memory requirement.
This highlights the impact and importance of the design choices and dataflow optimizations.

\niparagraph{Case study on protein sequencing.}
Long sentence protein sequencing can model protein interaction networks without the large sequence alignments~\cite{performer}. We follow the same methodology as in Performer~\cite{performer} on the TrEMBL protein sequencing dataset~\cite{uniprot2019uniprot} on a target accelerator with 16GB memory (\autoref{sec:flat_gpu}). We show the memory requirement when the number of attention blocks varies with the sequence lengths of 8K (\autoref{table:trembl_8k}) and 16K (\autoref{table:trembl_16k}). In both cases, FLAT unlocks the opportunity to use larger transformer models (\autoref{table:trembl_8k}) and/or larger sequence length (\autoref{table:trembl_16k}), potentially increasing model performance.
In summary, \dataflow outperforms recent DNN dataflow approaches~\cite{mindmapping, gamma, maestro, timeloop, cosa, eyeriss_isscc, tpuv3, nvdla, du2015shidiannao} on attention models owing to its specialized fusion and tiling tailored for attentions, enabling longer sequence lengths.

\section{Evaluation II: Accelerator Comparison}
\label{sec:exp_accel}
In this section, we contrast the performance of specific accelerator design points with and without the \dataflow dataflow.

\subsection{\rev{Cloud and Edge Accelerators}}
\label{sec:exp_accel_cloud_edge}
\begin{table}[!t]
\centering
\caption{\label{table:trembl_8k}\dataflow on the TrEMBL protein sequencing dataset~\cite{uniprot2019uniprot} on a target accelerator (Tesla T4 GPU~\cite{teslaT4}) with 16GB memory, using sequence length = 8K.}
\scriptsize
\resizebox{0.48\textwidth}{!}{
\renewcommand{\aboverulesep}{0pt}
\renewcommand{\belowrulesep}{0pt}
\begin{tabular}{l|ccccccc}\toprule
\multirow{2}{*}{\makecell{Memory \\ Req.(GB)}} &\multicolumn{6}{c}{Number of Attention Blocks (Sequence Length=8K)} \\\cmidrule{2-7}
&1 &2 &3 &4 &5 &6 \\\midrule
Baseline &4.6 &9.1 &13.7 &\makecell{18.2 \\ -OOM} &\makecell{22.2 \\ -OOM} &\makecell{27.3 \\ -OOM} \\\midrule
\dataflow &0.9 &1.8 &2.7 &3.5 &4.4 &5.3 \\
\bottomrule
\end{tabular}
}
\end{table}
We start by selecting two specific hardware design points, namely a Cloud and an Edge accelerator, with headline HW resources that closely resemble a TPU-v3~\cite{tpuv3} and Edge TPU~\cite{edgetpu,yazdanbakhsh2021evaluation}, as shown in \autoref{table:df_accel_config}.
We fix the on-chip buffer capacity to 512KB~\cite{edgetpu} and 32MB~\cite{tpuv3} for Edge and Cloud accelerators, respectively.
Analyzing these accelerators across different dataflow spaces (\autoref{table:df_config}), namely \naive, \flex, and \dataflow, forms a concrete and reasonably realistic accelerator design space.
Similar to previous sections, we name the optimal accelerator design point in each design space: \naive\-Edge, \flex-Opt-Edge, and \dataflow-Opt-Edge for edge accelerator, and \naive-Cloud, \flex-Opt-Cloud, and \dataflow-Opt-Cloud for cloud accelerator, respectively.
\begin{table}[!t]
\centering
\caption{\label{table:trembl_16k}\dataflow on the TrEMBL protein sequencing dataset~\cite{uniprot2019uniprot} on a target accelerator (Tesla T4 GPU~\cite{teslaT4}) with 16GB memory, using sequence length = 16K.}
\scriptsize
\resizebox{0.48\textwidth}{!}{
\renewcommand{\aboverulesep}{0pt}
\renewcommand{\belowrulesep}{0pt}
\begin{tabular}{l|ccccccc}\toprule
\multirow{2}{*}{\makecell{Memory \\ Req.(GB)}} &\multicolumn{6}{c}{Number of Attention Blocks (Sequence Length=16K)} \\\cmidrule{2-7}
&1 &2 &3 &4 &5 &6 \\\midrule
Baseline &\makecell{{17.5} \\ {-OOM}} &\makecell{35 \\ -OOM} &\makecell{52.5 \\ -OOM} &\makecell{70.0 \\ -OOM} &\makecell{87.5 \\ {-OOM}} &\makecell{{105.0} \\ {-OOM}} \\\midrule
\dataflow &2.8 &5.6 &8.5 &11.3 &14.1 &\makecell{{16.9} \\ {-OOM}} \\
\bottomrule
\end{tabular}
}
\end{table}

\niparagraph{Accelerator performance.}
As show in \autoref{fig:latency_ratio_a}, \flex-Opt-Edge and \dataflow-Opt-Edge share the same normalized runtime for K/Q/V/O and FF1/FF2. 
This similarity in performance is because in \dataflow-Opt-Edge, both K/Q/V/O and FF1/FF2 are treated as non-fused operators, and hence the map space for them are the same as the one in \flex-Opt-Edge. 
In edge accelerator, when the sequence length is 512, \dataflow-Opt-Edge and \flex-Opt-Edge both reach a near optimal performance. 
However, when the sequence length increases to 4K, 16K, and 64K, the performance gap between \dataflow-Opt-Edge and \flex-Opt-Edge widen.
For example, at sequence length of 64K, \dataflow-Opt-Edge runs 2.8$\times$ faster than \flex-Opt-Edge, showing the efficiency of our dataflow optimization.
In the cloud accelerator (\autoref{fig:latency_ratio}(b)), the performance difference between \dataflow-Opt-Cloud and \flex-Opt-Cloud exaggerates even further. For example, at sequence length of 64K, \dataflow-Opt-Cloud runs 3.07$\times$ faster than \flex-Opt-Cloud.
That is partly because of the larger model size for the cloud accelerator that enables \dataflow-Opt-Cloud to better utilize the on-chip hardware resources.

\niparagraph{Comparisons across different models.} 
\autoref{fig:exp_accel} compares the performance of different dataflow optimizations across various transformer models.
Compared to \flex-Opt-Edge, \dataflow-Opt-Edge delivers 1.75$\times$ speedup in edge accelerator, while significantly reducing the energy consumption by 44$\%$. 
In cloud accelerator, \dataflow-Opt-Cloud achieves 1.65$\times$ speedup and 55$\%$ energy savings over \flex-Opt-Cloud.
These results show the broad application of \dataflow in improving the performance of various attention-based models under different design constraints.

\niparagraph{Memory bandwidth requirement.}
Effectively using a limited off-chip bandwidth is an critical factor in the scalability of the hardware accelerator.
That is because most DNN operations are often memory-bound and the off-chip memory bandwidth is often shared across different microarchitectural components in the system.
In \autoref{fig:exp_accel_fig}, we show the peak off-chip bandwidth requirement to achieve a compute utilization over 0.95 for \lshort and \ashort attention operators. 
The left hand side of the \textit{U}-shape of \autoref{fig:exp_accel_fig} comes from the increase in the operational intensity and thus decrease of the bandwidth-boundedness as sequence length increases (\autoref{sec:challenge_opintensity}).

The right hand side of the \textit{U}-shape of \autoref{fig:exp_accel_fig} is caused by the quadratic and linear increase of on-chip memory requirement as sequence length increases for \flex and \dataflow, respectively.
On average, \dataflow-Opt-Cloud reduces the off-chip bandwidth requirement by 82$\%$ against \flex-Opt-Cloud.
Similarly, when evaluated under the edge scenario running \bench{BERT}, \dataflow-Opt-Edge achieves 71$\%$ reduction, on average, in the off-chip bandwidth requirement against \flex-Opt-Edge.
In summary, we demonstrate that \dataflow with its advantage of lowering on-chip buffer footprint can improve attention performance under existing Edge~\cite{edgetpu} and Cloud~\cite{tpuv3} DNN accelerators.

\subsection{\rev{\dataflow Compatibility with other Accelerators}}
\label{sec:exp_accel_gpu_attn}

\niparagraph{\dataflow compatibility on GPU.}
We implement and evaluate \dataflow on Nvidia-Tesla-T4~\cite{teslaT4} with 16GB memory (\autoref{sec:flat_gpu}). We use the BERT-Edge model and perform two experiments: (1) We fix the sequence length to 256 and sweep the batch size (\autoref{table:gpu_sweep_batch}),  and (2) we fix the batch size to one and sweep the sequence length (\autoref{table:gpu_sweep_seq}). \autoref{table:gpu_sweep_batch} shows that FLAT can run faster than baseline and supports larger batch sizes, whereas \autoref{table:gpu_sweep_seq} demonstrates that FLAT runs faster than baseline and supports up to 64K-word under the GPU-device we use.
\begin{table}[!t]\centering
\vspace{-0.2cm}
\caption{\rev{Runtime improvement of attention layer on Tesla-T4~\cite{teslaT4} GPU under different batch sizes.}} 
\scriptsize
\renewcommand{\aboverulesep}{0pt}
\renewcommand{\belowrulesep}{0pt}
\begin{tabular}{l|rrrrrrrr}\toprule
\multirow{2}{*}{\makecell{Runtime \\ (ms)}} &\multicolumn{7}{c}{Batch Size (Sequence Length=256)} \\\cmidrule{2-8}
&1 &16 &64 &128 &256 &1K &2K \\\midrule
Baseline &36 &630 &2,520 &5,230 &OOM &OOM &OOM \\\midrule
FLAT &28 &480 &1,870 &3,740 &7,560 &34,010 &OOM \\
\bottomrule
\end{tabular}
\label{table:gpu_sweep_batch}
\end{table}
\begin{table}[!t]\centering
\caption{\rev{Runtime improvement of attention layer on Tesla-T4~\cite{teslaT4} GPU under different sequence length.}}
\scriptsize
\renewcommand{\aboverulesep}{0pt}
\renewcommand{\belowrulesep}{0pt}
\begin{tabular}{l|rrrrrrrr}\toprule
\multirow{2}{*}{\makecell{Runtime \\ (ms)}} &\multicolumn{7}{c}{Sequence Length (Batch Size=1)} \\\cmidrule{2-8}
&128 &512 &2K &4K &16K &64K &128K \\\midrule
Baseline &12 &74 &697 &OOM &OOM &OOM &OOM \\\midrule
FLAT &11 &43 &175 &424 &4,599 &64,350 &OOM \\
\bottomrule
\end{tabular}
\label{table:gpu_sweep_seq}
\end{table}

\begin{figure}
\begin{center}
\includegraphics[width=0.99\linewidth]{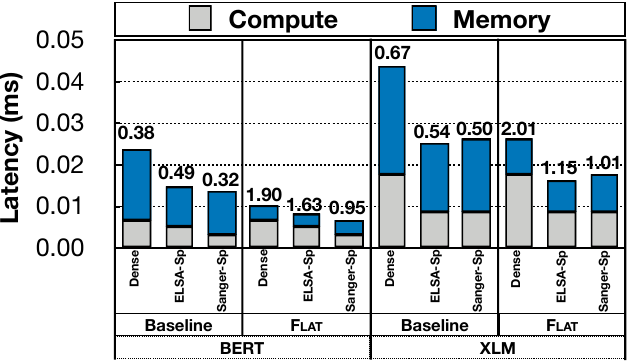}
\end{center}
\caption{\rev{The compute and memory time of L-A operations in BERT and XLM model under TPU-v3 configurations~\cite{tpuv3}. The valu on top of each bar indicates Compute-/Memory-time ratio. Compute-/Memory-time ratio (\textit{C/M}) < 1 means memory-boundedness; Likewise, \textit{C/M} > 1 means compute-boundedness, and the dataflow/system has more flexibility to operate to fully utilize the compute resources. We compare dense attention (Dense), ELSA-style sparse attention (ELSA-sp)~\cite{elsa}, Sanger-style sparse attention (Sanger-sp)~\cite{lu2021sanger} with and without \dataflow optimization. It shows that both ELSA~\cite{elsa} and Sanger~\cite{lu2021sanger} can effectively improve the attention performance, and both ELSA and Sanger can leverage \dataflow to reduce the memory-boundedness (note that \textit{C/M} increases to near or above 1.0 after applying \dataflow) and further improve the performance.}}
\label{fig:elsa_cpr}
\end{figure}

\niparagraph{\dataflow compatibility with sparse-attention accelerators.}
Recent sparse-attention accelerators such as ELSA~\cite{elsa} and Sanger~\cite{lu2021sanger} eschew model accuracy for improved performance.
Whereas \dataflow is a software-only method with no repercussions for model accuracy, which can readily be adapted across various platforms (e.g. Edge or Cloud). 
ELSA \textit{implicitly} supports and implements a limited form of inter-layer fusion operation at row granularity in hardware.
Sanger, a sparse-attention accelerator with a systolic array core, also employs one narrow form of fusion per row of PEs.
Both fusions are similar to one of many supported fusion granularity in \dataflow (\autoref{sec:constraint_dependency}).
In particular, \dataflow selects the proper granularity level (e.g. row, head, and batch as described in \autoref{subsec:granularity}) according to on-chip memory resources.
To quantitatively evaluate the impact of sparsity on \dataflow, we extended our methodology (\autoref{sec:accelerator_models}) to measure the memory-boundedness of L-A operators when L and A are \textit{sparse}.
We randomly sparsify L-A matrix with pruning ratio 50$\%$ in \bench{BERT} and \bench{XLM} on TPU-v3 platform resources~\cite{tpuv3}.
The results (\autoref{fig:elsa_cpr}) emphasize that even with a high degree of sparsity, these operations are still memory-bound, which warrants the benefits of \dataflow.

\section{Related Works}
\label{sec:relatedworks}
\vspace{-1mm}
\niparagraph{Dataflow and mapping.}
Most work on DNN hardware dataflow techniques focus on individual \CONV~\cite{nvdla, du2015shidiannao, chen2016eyeriss, gamma, timeloop, simba,lu2017flexflow, flexflow, yang2020interstellar, dmazerunner,gao2019tangram,tetris,deeptools,zhang2015optimizing, scaledeep,hypar,yazdanbakhsh2018ganax}, \GEMM~\cite{tpu,systolic_mapping,prime} operators, or loop reordering for transformer operations~\cite{markus}.
Some recent works consider fusion of multiple \CONV operator~\cite{fused_cnn,wei2018tgpa}.
Andrei et al.~\cite{ivanov2020data} merely targets operation fusion between MatMul operators and element-wise operators.
Fusing multiple heads of the attention operators~\cite{faster_transformer, nvidia_fusedmultihead_blog} primarily involves adding an additional loop over the H \emph{independent} heads, which is already captured by \flex.
They, however, do not explore dependent MatMul-Softmax-MatMul fusion, which is more complicated.
\dataflow targets such fusion and enables significantly higher performance.  

\niparagraph{Algorithmic optimization.}
Techniques such as quantization~\cite{shen2020q,kim2021bert,zafrir2019q8bert, zhang2020ternarybert}, pruning~\cite{wang2021spatten,leopard,sprint:micro,guo2019reweighted,wang2019structured,sajjad2020poor}, and distillation~\cite{sanh2019distilbert,jiao2019tinybert,sun2020mobilebert,wang2020minilm} are used for compressing Attention-based models.
There are a large body of algorithmic changes to attention mechanism~\cite{shen2018bi,parmar2018image,qiu2019blockwise,child2019generating,beltagy2020longformer}, learned sparsity~\cite{correia2019adaptively,kitaev2020reformer,roy2021efficient,tay2020synthesizer} low-rank and kernel methods~\cite{wang2020linformer,katharopoulos2020transformers,choromanski2020masked,performer}, and others~\cite{TrXL,beltagy2020longformer,compressive_transformer}. These techniques impact model quality and are orthogonal to the ideas developed in this paper. \dataflow can be leveraged in association with these techniques when deployed on DNN accelerators to further improve run time and energy.

\niparagraph{Matrix-Matrix fusion accelerators.} \rev{The core of Graph Neural Networks (GNNs) includes two consecutive matrix computations (``aggregate'' and ``combine''). GCNAX~\cite{li2021gcnax} and GRIP~\cite{kiningham2020grip} form a matrix-matrix loop fusion dataflow to optimize the throughput and energy efficiency. 
There are different challenge and focus for matrix-matrix fusion in GNNs and attentions.
1) The dataflows of GNN accelerators~\cite{li2021gcnax, kiningham2020grip} optimize matrix-matrix fusion (\autoref{fig:classify_table}(c)-left column), whereas the dataflows of attention optimizes matrix-Softmax-matrix fusion (\autoref{fig:classify_table}(c)-right column).
2) GNN includes one activation-weight and one activation-activation matrix computation, whereas attention has both matrix computations as activation-activation.
3) The key challenge of GNN is the sparsity in the ``aggreate'' matrix, whereas attention is challenged by the quadratic complexity of intermediate activation matrix between two matrix-multiplies.
}

\niparagraph{Attention accelerators.}
$A^{3}$~\cite{a3} and ELSA~\cite{elsa} propose dedicated attention accelerators and leverage approximate computation to accelerate attention layers.
\rev{Sanger~\cite{lu2021sanger} uses quantized query and key to predict attention matrix and rearranges the sparse attention matrix for better utilization. These technique trade-off performance with model quality.} \dataflow, by contrast, does not impact model quality, and is a generic yet efficient dataflow technique that can be leveraged on most existing  accelerators. \rev{SM6~\cite{tambe2021sm6} is an attention accelerator for RNN-based networks, which exposes different challenge and is orthogonal to this work.} 

\niparagraph{Compiler optimizations.}
Fusion is a classic compiler technique~\cite{wolfe1982optimizing,allen1992vector,gao1992collective,ding2004improving,kennedy1993maximizing,autotvm, baghdadi2019tiramisu, TACO, tensorflow_xla}. However, ML compilers employ fusion in a limited fashion to fuse matrix operators with element-wise operators~\cite{dnnfusion}.

\section{Conclusion}
\label{sec:conclusion}
We identify that running attention-based models with long sequences is challenging because of low reuse in certain attention operators and quadratic growth of intermediate memory footprint, both of which compound memory bandwidth requirements. We propose \dataflow, a novel dataflow for attention layers employing inter-operator fusion (the first work to investigate this for attention layers), interleaved execution, and efficient tiling to
enhance the operational intensity and provide high compute utilization, reduced off-chip bandwidth requirements and scalability to long sequence lengths.

\section*{Acknowledgment}
We would like to extend our gratitude towards Parthasarathy Ranganathan, Nishant Patil, James Laudon, Stella Aslibekyan, and extended Google Research, Brain Team for their invaluable feedback and comments. We also thank Prasanth Chatarasi for feedback on early drafts. This work was supported in-part by NSF Award \#1909900.
\section*{}

\bibliographystyle{plain}
\bibliography{main}

\end{document}